# On the Integration of Physics-Based Machine Learning with Hierarchical Bayesian Modeling Techniques

Omid Sedehi[1], Antonina M. Kosikova[2], Costas Papadimitriou[3], Lambros S. Katafygiotis[4]

**Abstract**
Machine Learning (ML) has widely been used for modeling and predicting physical systems. These techniques offer high expressive power and good generalizability for interpolation within observed data sets. However, the disadvantage of black-box models is that they underperform under blind conditions since no physical knowledge is incorporated. Physics-based ML aims to address this problem by retaining the mathematical flexibility of ML techniques while incorporating physics. In accord, this paper proposes to embed mechanics-based models into the mean function of a Gaussian Process (GP) model and characterize potential discrepancies through kernel machines. A specific class of kernel function is promoted, which has a connection with the gradient of the physics-based model with respect to the input and parameters and shares similarity with the exact Auto-covariance function of linear dynamical systems. The spectral properties of the kernel function enable considering dominant periodic processes originating from physics misspecification. Nevertheless, the stationarity of the kernel function is a difficult hurdle in the sequential processing of long data sets, resolved through hierarchical Bayesian techniques. This implementation is also advantageous to mitigate computational costs, alleviating the scalability of GPs when dealing with sequential data. Using numerical and experimental examples, potential applications of the proposed method to structural dynamics inverse problems are demonstrated.

**Abbreviations**

| | |
|---|---|
| BIC | Bayesian Information Criterion |
| ML | Machine Learning |
| GP | Gaussian Process |
| GWN | Gaussian White Noise |
| MMTE | Multi-Modal Trigonometric Exponential |
| MPV | Most Probable Value |
| NN | Neural Network |
| PSD | Power Spectral Density |
| SHM | Structural Health Monitoring |
| SVD | Singular Value Decomposition |
| RMS | Root-Mean-Square |

**Symbols**

| | |
|---|---|
| $\alpha$ | Damping coefficient |

[1] Postdoctoral Fellow, Department of Civil and Environmental Engineering, The Hong Kong University of Science and Technology, Hong Kong, Email: osedehi@connect.ust.hk
[2] Ph.D. Student, Department of Civil and Environmental Engineering, The Hong Kong University of Science and Technology, Hong Kong, Email: akosikova@connect.ust.hk
[3] Professor, Department of Mechanical Engineering, University of Thessaly, Volos, Greece, Email: costasp@uth.gr
[4] Professor, Department of Civil and Environmental Engineering, The Hong Kong University of Science and Technology, Hong Kong, Email: katafygiotis.lambros@gmail.com

| | |
|---|---|
| $\beta$ | Damping coefficient |
| $\delta(.)$ | Dirac delta function |
| $\varphi(.)$ | Latent operator of Gaussian Process |
| $\hat{\varpi}$ | Modal frequency of structure |
| $\sigma_k^2$ | Amplitude of the cosine kernel function |
| $\sigma_n^2$ | Noise variance |
| $\tau$ | Time difference |
| $\omega$ | Frequency |
| $\omega_k$ | Frequency of the cosine kernel function |
| $d(\zeta,\zeta')$ | Euclidean distance of $\zeta$ and $\zeta'$ |
| $f(\mathbf{x},\boldsymbol{\theta})$ | Simulated response of physics-based model $M_\theta$ |
| $k_\varepsilon()$ | Kernel covariance function |
| $k_\theta()$ | Physics-based tangent kernel function |
| $\ell_k$ | Correlation length |
| $m$ | Number of cosine functions in the MMTE kernel |
| $M_\theta$ | Physics-based model with parameters $\boldsymbol{\theta}$ |
| $S_\varepsilon(\omega;\boldsymbol{\phi})$ | PSD of MMTE kernel function |
| $\boldsymbol{\delta}$ | Hyper-parameter vector |
| $\boldsymbol{\varepsilon}_{\boldsymbol{\delta}_i}$ | Prediction error process |
| $\boldsymbol{\phi}$ | Parameters of the MMTE kernel |
| $\boldsymbol{\zeta}$ | Auxiliary variables |
| $\boldsymbol{\mu}_{i|i-1}$ | Conditional mean vector |
| $\boldsymbol{\mu}_\theta$ | Hyper mean of $\boldsymbol{\theta}$ |
| $\boldsymbol{\Sigma}_\theta$ | Hyper covariance matrix of $\boldsymbol{\theta}$ |
| $\boldsymbol{\Sigma}_{i|i-1}$ | Conditional covariance matrix |
| $\boldsymbol{\theta}$ | Physics-based model parameters |
| $\mathbf{C}$ | Damping matrix |
| $\mathbf{J}$ | Gradient matrix |
| $\mathbf{K}$ | Stiffness matrix |
| $\mathbf{K}(\boldsymbol{\phi})$ | Full covariance matrix with parameters $\boldsymbol{\phi}$ |
| $\mathbf{M}$ | Mass matrix |
| $\mathbf{Q}_\delta$ | Random-walk model parameters |
| $\mathbf{x}$ | Initial and input conditions |
| $\mathbf{y}$ | Measured time-history response |

## 1. Introduction

First principle models of physical systems shape the backbone of computational science and engineering, enormously developed for a wide range of real-life applications. The performance of these models directly depends on the developers' knowledge and expertise in how accurately they can describe the physical system of interest. Therefore, validation and verification of physics-based models have been explored continuously as a necessary and active task by investigators of different scientific and engineering disciplines. Although traditional approaches to data-driven model identification exist, there has been a critical need for proposing systematic tools with enhanced robustness and accuracy [1,2].

Physics-based Machine Learning (ML) has recently emerged in computational physics and mechanics through deep Neural Networks (NNs) and Gaussian Processes (GPs) [3,4]. These models might deem completely different from each other, but recent works reveal that there are interrelations and connections as well [5,6]. In terms of performance, it is notable that deep networks enjoy moderate generalizability and high expressive power. Nonetheless, uncertainties are often intractable owing to complex network architecture. Moreover, deep networks do not offer physical interpretability and demand high computational costs, especially when employed in conjunction with detailed physics-based models. Conversely, GP models can embed physics-based models in their mean process and express correlation and uncertainty through kernel covariance machines. In this respect, physics-based models and kernel machines can be parameterized through a common set of uncertain parameters, whose identification can be addressed through maximum likelihood estimation and Bayesian inference [7,8].

Physics-informed GPs are highly advantageous as they allow compensate for the discrepancy between the physics-based models and measured data [9]. In this context, the model discrepancy might originate from the misspecification of the governing physical equations, initial conditions, boundary equations, and physical parameters. This type of uncertainty can be captured through GP models only if a proper kernel covariance function is established [10]. In the ML literature, common choices of kernel machines include Exponential, Exponential-Sinusoidal, Polynomial, Matern', Neural, and Rational covariance functions [7]. Such kernels can also be combined through basic algebraic operations, e.g., summation and multiplication, to simulate relationships that are more complex [7,11]. These kernel functions are founded upon the stationarity of the underlying stochastic processes, which describe the correlation between the data points based on a measure of proximity of their projection onto an auxiliary latent space [12]. Such an auxiliary space of parameters can be specified using available input information although it can be possible to infer them from the data as well. In any case, the stationarity of kernel functions allows the Wiener-Khintchine theorem to provide a spectral realization, which helps better understand theoretical properties and guides users through the selection process [7].

At first glance, it might deem reasonable to incorporate the above covariance functions for identifying physics-based models. For instance, GP models have customarily been associated with squared-exponential kernel function, correlating data points based on the Euclidean distance of the auxiliary input space [13–15]. In most practical applications, this class of GPs, known as kriging, can perform well for interpolation within training sets of measured data. Nevertheless, when it comes to predicting held-out data sets, they provide disproportionate predictions, which do not match dominant trends appearing in the model discrepancy [16]. This issue should be searched in the fact that such kernel machines do not benefit from any physical knowledge to capture the prevailing correlation patterns and dependencies. On top of this problem, the model discrepancy is unlikely to be wide-sense stationary as they can vary across time and space domains, whereas the foregoing kernel functions tie well within stationarity. Therefore, there is a research gap in kernel-based GP solutions, which incorporate contextual physical knowledge of specific applications and consider non-stationarity.

Despite these shortcomings, GP models have been successful in the context of Structural Health Monitoring (SHM) [17–19]. Yuen and Katafygiotis have used GP models to describe the correlation of prediction errors in Bayesian modal updating problems comparing the exact covariance matrix with the sample autocorrelation function [20,21]. Papadimitriou and Lombaert have used GPs with exponential kernel functions for considering the spatial correlation of closely spaced sensors and the redundancy of sensing information to avoid clustering of sensors at certain locations [22]. Simoen et al. have examined numerous classes of kernel functions and established a Bayesian approach to trade-off accuracy with the complexity of kernel functions [10]. Kosikova et al. propose a new model selection scheme, which also incorporates the quality of predictions for kernel selection [16]. Avendano-Valencia et al. have used GP models for describing the temporal variation of dynamical properties with operational and environmental conditions [23]. In the same vein, Zhu and Au have employed GPs to describe the relationship of modal properties with loadings and prediction errors spectral properties [24]. The state-space representation of GPs has been used for describing and identifying unknown input forces [25,26]. Jiang et al. have applied GP-based models for updating the FE models of Miter Gates [27]. Gardner et al. have applied kernel-based domain adaptation for population-based SHM [28]. Ramancha et al. have identified the exact kernel covariance function for the identification of linear structures [29]. Cross et al. have used Matern' covariance functions to create geometry-consistent GPs for describing acoustic emission mappings [30]. This on-going trend seems to continue in the next years as new inspirations and applications step into the research spotlight.

This study expands upon physics-based ML by pushing forward a new class of kernel covariance functions and considering non-stationary effects. The kernel function is investigated theoretically, shown to be consisting of a Multi-Modal Trigonometric Exponential (MMTE) covariance function and a physics-driven tangent covariance structure. By doing so, the gradient information of physics-based models is considered in the covariance function, offering robustness when facing parametric misspecification. Non-stationary effects are considered through hierarchical Bayesian modeling techniques, e.g., [31–34], where the temporal variation of parameters is considered through partitioning the measured data into sub-partitions across which stationary assumptions remain valid. Once the theoretical findings are explained, numerical and experimental examples are provided to highlight potential applications and merits compared to the classical Bayesian approach.

This paper continues with Section 2, which formalizes the new formulation and introduces the new kernel function. Section 3 investigates the theoretical properties of the proposed framework, and Section 4 proposes a computational algorithm to summarize the detailed flow of steps. Section 5 demonstrates the proposed methodology through numerical and experimental examples, and Section 6 draws conclusions and maps our future works.

## 2. Proposed Methodology
### 2.1. Physics-informed Gaussian Process

Let $\mathbf{y} \in \mathbb{R}^{nN_o}$ denote the measured time-history response of a mechanical system subjected to known initial and input conditions $\mathbf{x} \in \mathbb{R}^{nN_x}$, where $N_o$ is the number of measured quantities, $N_x$ is the dimension of the input space, and $n$ is the number of time increments. The objective is to carry out a Supervised Learning Task and update a physics-based model $M_\theta$ and its unknown parameters $\boldsymbol{\theta} \in \mathbb{R}^{N_\theta}$ from the input-output pairs $(\mathbf{x}, \mathbf{y})$. For this purpose, we first define that, given the input $\mathbf{x}$ and the parameter $\boldsymbol{\theta}$, the physics-based model $M_\theta$ provides an explicit functional relationship to generate the physical responses as follows:

$$M_\theta : \begin{cases} \mathbb{R}^{nN_x + N_\theta} \mapsto \mathbb{R}^{N_o} \\ (\mathbf{x}, \boldsymbol{\theta}) \mapsto f(\mathbf{x}, \boldsymbol{\theta}) \end{cases} \quad (1)$$

where $f(\mathbf{x}, \boldsymbol{\theta}) \in \mathbb{R}^{N_o}$ is the simulated response of the physics-based model $M_\theta$. Due to modeling errors and measurement noise, the model responses deviate from the measured ones. Additionally, the choice of structural model parameters $\boldsymbol{\theta}$ can considerably affect simulated responses. To overcome these issues, a class of physics-informed kernel machines is employed to describe the probabilistic relationship between the output $\mathbf{y}$ and the input $\mathbf{x}$ as

$$\mathbf{y} \mid \mathbf{x}, \boldsymbol{\theta}, \boldsymbol{\phi} \sim GP\left(\mathbf{y} \mid \underbrace{f(\mathbf{x}, \boldsymbol{\theta})}_{Physics-based\ Model}, \underbrace{k_\varepsilon(\boldsymbol{\zeta}, \boldsymbol{\zeta}'; \boldsymbol{\phi})}_{Kernel\ Machines}\right) \quad (2)$$

where $GP(.\mid m, s)$ is a GP with $m$ mean and $s$ covariance function; $k_\varepsilon(\boldsymbol{\zeta}, \boldsymbol{\zeta}'; \boldsymbol{\phi})$ is a kernel machine described via the unknown parameters $\boldsymbol{\phi} \in \mathbb{R}^{N_\phi}$ and the auxiliary variables $\boldsymbol{\zeta} \in \mathbb{R}^{N_\zeta}$. The kernel $k_\varepsilon(\zeta,\zeta';\boldsymbol{\phi})$ characterizes a latent dot-product Euclidean space, which projects data points into the multi-dimensional space of the auxiliary parameters $\zeta \in \mathbb{R}^{N_\zeta}$ and describes the correlation between data points based on the proximity of the auxiliary variables. This specification essentially implies that the kernel function can be written as

$$k_\varepsilon : \begin{cases} \mathbb{R}^{N_\zeta \times N_\zeta} \mapsto \mathbb{R} \\ (\zeta,\zeta') \mapsto k_\varepsilon(\zeta,\zeta';\boldsymbol{\phi}) = \left(\varphi(\zeta)\right)^T \varphi(\zeta') \end{cases} \quad (3)$$

where $\varphi(.) : \mathbb{R}^{N_\zeta} \mapsto \mathbb{R}^{N_\varphi}$ is a latent operator acting upon $\zeta$'s to describe the correlation between data points. It should be noted that, in this paper, the mathematical expression of $k_\varepsilon(\zeta,\zeta';\boldsymbol{\phi})$ is assumed to be available, so there is no need to acquire or specify the operator $\varphi(.)$ explicitly.

Based on Eq. (2), the likelihood function of an observed vector like $\mathbf{y}$ can be described as

$$p(\mathbf{y}|\mathbf{x},\boldsymbol{\theta},\boldsymbol{\phi}) = N\left(\mathbf{y}|f(\mathbf{x},\boldsymbol{\theta}),\mathbf{K}(\boldsymbol{\phi})\right) \quad (4)$$

where $\mathbf{K}(\boldsymbol{\phi}) \in \mathbb{R}^{nN_o \times nN_o}$ is the prediction error covariance matrix parameterized by known parameters $\boldsymbol{\phi}$, which will be specified later.

## 2.2. Tangent Kernel Machines

The physical model $M_\theta$ provides a one-to-one relationship between $\boldsymbol{\theta}$ and $\mathbf{y}$, when $\mathbf{x}$ is known. Due to modeling errors, any choice of $\boldsymbol{\theta}$ can create some residual mismatch between the model and measured responses. This type of uncertainty is mixed with other sources of errors but can be modelled by releasing $\boldsymbol{\theta}$ to vary according to a parameterized Gaussian distribution, given as

$$\boldsymbol{\theta}|\boldsymbol{\mu_\theta},\boldsymbol{\Sigma_\theta} \sim N\left(\boldsymbol{\theta}|\boldsymbol{\mu_\theta},\boldsymbol{\Sigma_\theta}\right) \quad (5)$$

where $\boldsymbol{\mu_\theta} \in \mathbb{R}^{N_\theta}$ and $\boldsymbol{\Sigma_\theta} \in \mathbb{R}^{N_\theta \times N_\theta}$ are the hyper mean and covariance of the physics-based model parameters. These hyper-parameters $\{\boldsymbol{\mu_\theta},\boldsymbol{\Sigma_\theta}\}$ should also be inferred from the data. For this purpose, the Bayes' rule is used to provide the posterior distribution of all unknown parameters as

$$\begin{aligned} p(\boldsymbol{\phi},\boldsymbol{\theta},\boldsymbol{\mu_\theta},\boldsymbol{\Sigma_\theta}|\mathbf{x},\mathbf{y}) &\propto p(\mathbf{y}|\mathbf{x},\boldsymbol{\phi},\boldsymbol{\theta},\boldsymbol{\mu_\theta},\boldsymbol{\Sigma_\theta})p(\boldsymbol{\phi},\boldsymbol{\theta},\boldsymbol{\mu_\theta},\boldsymbol{\Sigma_\theta}) \\ &\propto p(\mathbf{y}|\mathbf{x},\boldsymbol{\theta},\boldsymbol{\phi})p(\boldsymbol{\theta}|\boldsymbol{\mu_\theta},\boldsymbol{\Sigma_\theta})p(\boldsymbol{\mu_\theta},\boldsymbol{\Sigma_\theta})p(\boldsymbol{\phi}) \\ &\propto N(\mathbf{y}|f(\mathbf{x},\boldsymbol{\theta}),\mathbf{K}(\boldsymbol{\phi}))N(\boldsymbol{\theta}|\boldsymbol{\mu_\theta},\boldsymbol{\Sigma_\theta}) \end{aligned} \quad (6)$$

In this equation, the likelihood function $p(\mathbf{y}|\mathbf{x},\boldsymbol{\theta},\boldsymbol{\phi})$ is replaced from Eq. (4); the conditional prior distribution $p(\boldsymbol{\theta}|\boldsymbol{\mu_\theta},\boldsymbol{\Sigma_\theta})$ is replaced from Eq. (5); other prior distributions, i.e., $p(\boldsymbol{\mu_\theta},\boldsymbol{\Sigma_\theta})$ and $p(\boldsymbol{\phi})$ are considered uniform.

Although the posterior distribution in Eq. (6) is general, its calculation can be intractable due to the interdependence of parameters and the large number of parameters involved. Thus, from a computational standpoint, it is advantageous to marginalize some of the parameters as nuisance. For this purpose, we first approximate the functional relationship $f(\mathbf{x},\boldsymbol{\theta})$ of the model response through a first-order Taylor series expansion around $\forall \mathbf{x}$; $\boldsymbol{\theta} = \boldsymbol{\mu_\theta}$, which leads to

$$f(\mathbf{x},\boldsymbol{\theta}) \approx f(\mathbf{x},\boldsymbol{\mu_\theta}) + \mathbf{J}(\boldsymbol{\theta} - \boldsymbol{\mu_\theta}) \quad (7)$$

where $\mathbf{J} = \partial f(\mathbf{x},\boldsymbol{\theta})/\partial \boldsymbol{\theta}^T \big|_{\boldsymbol{\theta}=\boldsymbol{\mu_\theta}}$ is the gradient matrix evaluated at the expansion point. When this linearization is substituted into Eq. (7), it turns out that the model parameters $\boldsymbol{\theta}$ can be integrated out explicitly, leading to

$$p(\boldsymbol{\phi},\boldsymbol{\mu_\theta},\boldsymbol{\Sigma_\theta}|\mathbf{x},\mathbf{y}) \propto \int_{\boldsymbol{\theta}} N(\mathbf{y}|f(\mathbf{x},\boldsymbol{\theta}),\mathbf{K}(\boldsymbol{\phi}))N(\boldsymbol{\theta}|\boldsymbol{\mu_\theta},\boldsymbol{\Sigma_\theta})d\boldsymbol{\theta} \approx N\left(\mathbf{y}|f(\mathbf{x},\boldsymbol{\mu_\theta}),\mathbf{J}\boldsymbol{\Sigma_\theta}\mathbf{J}^T + \mathbf{K}(\boldsymbol{\phi})\right) \quad (8)$$

This result is a direct consequence of Theorem 1, discussed in Appendix (A), which simplifies the integral of two Gaussian distributions whose mean vectors are a linear function of the integral variable.

In this equation, the expression $\mathbf{J}\boldsymbol{\Sigma}_\theta \mathbf{J}^T$ is the tangent covariance matrix, incorporated through further parameterization of the physical parameters. When the matrix $\boldsymbol{\Sigma}_\theta$ is diagonal, the expression $\mathbf{J}\boldsymbol{\Sigma}_\theta \mathbf{J}^T$ can be simplified into

$$\mathbf{J}\boldsymbol{\Sigma}_\theta \mathbf{J}^T = \sum_{j=1}^{N_\theta} \sigma_{\theta_j}^2 \left[ \frac{\partial f(\mathbf{x},\boldsymbol{\theta})}{\partial \theta_j} \frac{\partial f^T(\mathbf{x},\boldsymbol{\theta})}{\partial \theta_j} \right]_{\boldsymbol{\theta}=\boldsymbol{\mu}_\theta} \tag{9}$$

where $\sigma_{\theta_j}^2$ denotes the $j^{th}$ diagonal element of $\boldsymbol{\Sigma}_\theta$. This covariance matrix characterizes the correlation between the data points through the following tangent kernel function:

$$k_\theta(x_p, x_q; \{\boldsymbol{\mu}_\theta, \boldsymbol{\Sigma}_\theta\}) = \sum_{j=1}^{N_\theta} \sigma_{\theta_j}^2 \left[ \frac{\partial f(x_p,\boldsymbol{\theta})}{\partial \theta_j} \frac{\partial f^T(x_q,\boldsymbol{\theta})}{\partial \theta_j} \right]_{\boldsymbol{\theta}=\boldsymbol{\mu}_\theta} \tag{10}$$

where $f(x_p, \boldsymbol{\theta})$ and $f(x_q, \boldsymbol{\theta})$ returns the physics-based model response when the physical system of interest is subjected to the incremental input $x_p$ and $x_q$; $k_\theta(x_p, x_q; \{\boldsymbol{\mu}_\theta, \boldsymbol{\Sigma}_\theta\})$ is the physics-based tangent kernel function, describing the correlation between the instances $x_p$ and $x_q$. On this basis, the following features of the tangent kernel function can be highlighted:

- The correlation between the data points $p$ and $q$ grows with the sensitivity of the physical response with respect to $\boldsymbol{\theta}$, given as $[\partial f(x_p,\boldsymbol{\theta})/\partial \theta_j]_{\boldsymbol{\theta}=\boldsymbol{\mu}_\theta}$ and $[\partial f(x_q,\boldsymbol{\theta})/\partial \theta_j]_{\boldsymbol{\theta}=\boldsymbol{\mu}_\theta}$.
- The tangent kernel $k_\theta(x_p, x_q; \{\boldsymbol{\mu}_\theta, \boldsymbol{\Sigma}_\theta\})$ only depends on the response sensitivities and the hyper-parameters. This implies that the variance of the $p^{th}$ data point is given proportional to the squared response sensitivity $[\partial f(x_p,\boldsymbol{\theta})/\partial \theta_j]^2_{\boldsymbol{\theta}=\boldsymbol{\mu}_\theta}$.
- Based on Eq. (10), the cross-correlation between the responses of different DOFs is proportional to the multiplication of their corresponding response sensitivities. This means that, for a fixed parameter covariance matrix $\boldsymbol{\Sigma}_\theta$, responses that are more sensitive yield larger uncertainty and vice versa.

**2.3. Trigonometric Kernel Machine**

In this paper, the MMTE kernel function, proposed in [16], is used for describing the correlation between data points. This kernel function reads as the mixture of squared-exponential and cosine functions, which is summed up with an isotropic variance to account for potential measurement noise, as given by

$$k_\varepsilon(\zeta, \zeta'; \boldsymbol{\phi}) = \left[ \sum_{k=1}^{m} \sigma_k^2 \exp\left(-\frac{d^2(\zeta,\zeta')}{\ell_k^2}\right) \cos\left(\omega_k d(\zeta,\zeta')\right) \right] + \sigma_n^2 \delta(d(\zeta,\zeta')) \tag{11}$$

where $\sigma_k^2$ is the amplitude of the correlation, $\sigma_n^2$ is the noise variance, $\ell_k$ is the correlation length of the $k^{th}$ exponential function, $\omega_k$ is the frequency of the $k^{th}$ cosine function, $d(\zeta,\zeta')$ is the Euclidean distance between $\zeta$ and $\zeta'$, and $\delta(.)$ is Dirac delta function. This specification of the kernel function implies that the kernel parameters can be collected into $\boldsymbol{\phi} = [\sigma_1^2 \ \ell_1^2 \ \omega_1 \ ... \ \sigma_m^2 \ \ell_m^2 \ \omega_m \ \sigma_n^2]^T$.

This definition implies that the covariance matrix $\mathbf{K}(\boldsymbol{\phi})$ embeds the elements $[k_{pq}] = k_\varepsilon(\zeta_p, \zeta_q; \boldsymbol{\phi})$. In this paper, the auxiliary parameters $\zeta_p$ and $\zeta_q$ are considered the corresponding time instants, and the distance measure $d(\zeta_p, \zeta_q)$ can reduce into the temporal distance $\tau = |t_q - t_p|$. Thus, the MMTE function can be rewritten as

$$k_\varepsilon(\tau;\boldsymbol{\phi}) = \left[\sum_{k=1}^{m}\sigma_{k,i}^2 \exp\left(-\frac{\tau^2}{\ell_k^2}\right)\cos(\omega_k\tau)\right] + \sigma_n^2\delta(\tau) \tag{12}$$

This particular form of covariance function shares similarity with the Auto-covariance function of linear dynamical systems under Gaussian White Noise (GWN) input [35]. It allows calculating the Fourier Transform of the above kernel function to arrive at its Power Spectral Density (PSD), which helps understand how the correlation is modeled. By doing so, the PSD of $k_\varepsilon(\tau;\boldsymbol{\phi})$ is obtained as

$$S_\varepsilon(\omega;\boldsymbol{\phi}) = \frac{1}{2}\sum_{k=1}^{m}\sqrt{\pi}\sigma_k^2\ell_k\left[\exp\left(-\frac{\ell_k^2}{4}(\omega+\omega_k)^2\right) + \exp\left(-\frac{\ell_k^2}{4}(\omega-\omega_k)^2\right)\right] + \sigma_n^2 \tag{13}$$

A mathematical proof for this expression is provided in Appendix (B). At this point, we consider a numerical example to visualize this kernel function in the frequency domain. For this purpose, let the parameters $m=2$, $\sigma_1^2=2$, $\sigma_2^2=8$, $\ell_1^2=5$, $\ell_2^2=2.5$, $\omega_1=2$, $\omega_2=10$, $\sigma_n^2=0.5$. Fig. 1(a) displays the corresponding kernel function, exhibiting how the MMTE kernel function composes two exponentially decaying cosines. Fig. 1(b) shows the corresponding PSD function given by Eq. (13), which indicates two peaks at 2 rad/s and 10 rad/s frequencies. Based on this demonstration, it can concluded that the MMTE can be suitable for modeling temporal correction when the prediction error process is composed of a few mixing dominant sinusoidal functions.

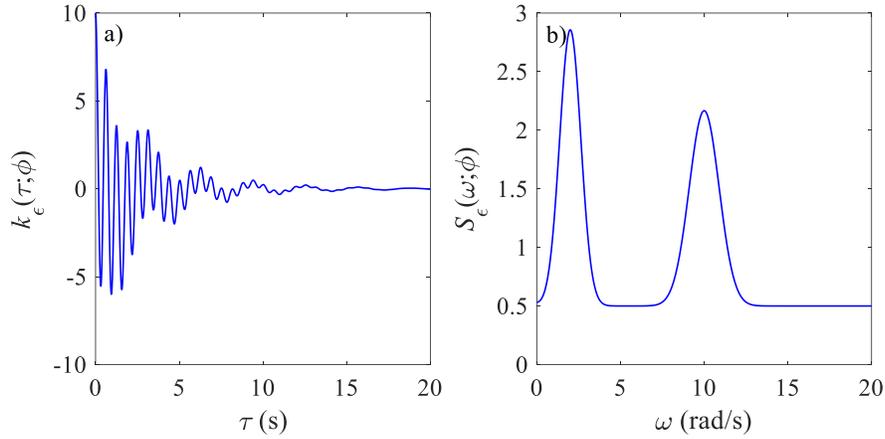

**Fig. 1.** (a) MMTE kernel covariance function (b) the corresponding PSD function

Despite desirable spectral properties of this kernel function, the underlying stationarity limits its application. This problem is tackled in the next section through hierarchical Bayesian techniques.

## 3. Hierarchical Bayesian Formulation
### 3.1. Joint Posterior Distribution

Let $\mathbf{X}=[\mathbf{x}_1^T\ \mathbf{x}_2^T\ ...\ \mathbf{x}_{N_D}^T]^T$ and $\mathbf{Y}=[\mathbf{y}_1^T\ \mathbf{y}_2^T\ ...\ \mathbf{y}_{N_D}^T]^T$ comprise a data set partitioned into $N_D$ segments, where $\mathbf{x}_i$ is known input and $\mathbf{y}_i$ is measured physical responses. The objective is the same as the pervious section, which is the Supervised Learning Task of inferring the unknown parameters of the physics-based model from this multi-partition data. The formulation presented earlier can similarly be applied to address this problem; however, when the posterior distribution in Eq. (8) is used for inferring the unknown parameters from each individual pair $(\mathbf{x}_i,\mathbf{y}_i)$, the unknown parameters and their distribution appear to vary across partitions. This variability is concerned with the parameters of the physics-based model $\boldsymbol{\theta}$, as well as the MMTE kernel parameters $\boldsymbol{\phi}$. Thus, the parameter vectors $\boldsymbol{\theta}_i$ and $\boldsymbol{\phi}_i$ are introduced through adding the subscript $i$, which represent the correspondence to the specific pair $(\mathbf{x}_i, \mathbf{y}_i)$. Additionally, the parameters $\{\boldsymbol{\theta}_i\}_{i=1}^{N_D}$ are considered mutually independent and parameterized by considering the following prior distribution in a similar sense as Eq. (5), leading to

$$\boldsymbol{\theta}_i \mid \boldsymbol{\mu}_{\boldsymbol{\theta}_i}, \boldsymbol{\Sigma}_{\boldsymbol{\theta}_i} \sim N(\boldsymbol{\theta}_i \mid \boldsymbol{\mu}_{\boldsymbol{\theta}_i}, \boldsymbol{\Sigma}_{\boldsymbol{\theta}_i}) \quad , \quad \forall i = \{1, ..., N_D\} \tag{14}$$

where $\boldsymbol{\mu}_{\boldsymbol{\theta}_i} \in \mathbb{R}^{N_\theta}$ and $\boldsymbol{\Sigma}_{\boldsymbol{\theta}_i} \in \mathbb{R}^{N_\theta \times N_\theta}$ are the mean vector and covariance matrix of the partition-specific parameters $\boldsymbol{\theta}_i$, respectively.

The parameters collected in $\boldsymbol{\delta}_i = vec\{\boldsymbol{\mu}_{\boldsymbol{\theta}_i}, \boldsymbol{\Sigma}_{\boldsymbol{\theta}_i}, \boldsymbol{\phi}_i\} \in \mathbb{R}^{N_\delta}$ are variable across data partitions. Due to this variability, knowledge transfer across the partitions is not possible unless additional assumptions are considered to regulate the variation of these parameters across partitions. For this purpose, the parameters $\boldsymbol{\delta}_i$'s are considered sequentially dependent, governed by a first-order random walk model, written as

$$\boldsymbol{\delta}_i = \boldsymbol{\delta}_{i-1} + \boldsymbol{\varepsilon}_{\boldsymbol{\delta}_i} \quad , \quad \forall i = \{1, ..., N_D\} \tag{15}$$

where $\boldsymbol{\varepsilon}_{\boldsymbol{\delta}_i} \in \mathbb{R}^{N_\delta}$ is a zero-mean GWN process with covariance matrix $\mathbf{Q}_{\boldsymbol{\delta}} = \text{block-diag}(\mathbf{Q}_{\boldsymbol{\mu}_\theta}, \mathbf{Q}_{\boldsymbol{\Sigma}_\theta}, \mathbf{Q}_{\boldsymbol{\phi}}) \in \mathbb{R}^{N_\delta \times N_\delta}$. Therefore, the conditional prior distribution of $\boldsymbol{\delta}_i$'s can be written as

$$\boldsymbol{\delta}_i \mid \boldsymbol{\delta}_{i-1}, \mathbf{Q}_{\boldsymbol{\delta}} \sim N(\boldsymbol{\delta}_i \mid \boldsymbol{\delta}_{i-1}, \mathbf{Q}_{\boldsymbol{\delta}}) \quad , \quad \forall i = \{1, ..., N_D\} \tag{16}$$

Based on Eqs. (14) and (16), the joint prior distribution turns out to be a hierarchical distribution, written as

$$p(\{\boldsymbol{\theta}_i, \boldsymbol{\delta}_i\}_{i=1}^{N_D}, \mathbf{Q}_{\boldsymbol{\delta}}) = \left[ \prod_{i=1}^{N_D} N(\boldsymbol{\theta}_i \mid \boldsymbol{\mu}_{\boldsymbol{\theta}_i}, \boldsymbol{\Sigma}_{\boldsymbol{\theta}_i}) N(\boldsymbol{\delta}_i \mid \boldsymbol{\delta}_{i-1}, \mathbf{Q}_{\boldsymbol{\delta}}) \right] p(\mathbf{Q}_{\boldsymbol{\delta}}) \tag{17}$$

where $p(\{\boldsymbol{\theta}_i, \boldsymbol{\delta}_i\}_{i=1}^{N_D}, \mathbf{Q}_{\boldsymbol{\delta}})$ is the joint prior distribution of all parameters, $p(\mathbf{Q}_{\boldsymbol{\delta}})$ is the prior distribution of the hyper-parameters $\mathbf{Q}_{\boldsymbol{\delta}}$, and $\boldsymbol{\delta}_0 = vec\{\boldsymbol{\mu}_{\boldsymbol{\theta}_0}, \boldsymbol{\Sigma}_{\boldsymbol{\theta}_0}, \boldsymbol{\phi}_0\} \in \mathbb{R}^{N_\delta}$ can be set to a prior estimate of the parameters. Having made this assumption, the Bayes' rule gives the joint posterior distribution of all parameters as follows:

$$p(\{\boldsymbol{\theta}_i, \boldsymbol{\delta}_i\}_{i=1}^{N_D}, \mathbf{Q}_{\boldsymbol{\delta}} \mid \mathbf{X}, \mathbf{Y}) \propto p(\mathbf{Y} \mid \{\boldsymbol{\theta}_i, \boldsymbol{\delta}_i\}_{i=1}^{N_D}, \mathbf{Q}_{\boldsymbol{\delta}}, \mathbf{X}) p(\{\boldsymbol{\theta}_i, \boldsymbol{\delta}_i\}_{i=1}^{N_D}, \mathbf{Q}_{\boldsymbol{\delta}}) \tag{18}$$

where the prior distribution $p(\{\boldsymbol{\theta}_i, \boldsymbol{\delta}_i\}_{i=1}^{N_D}, \mathbf{Q}_{\boldsymbol{\delta}})$ is given by Eq. (17), and $p(\mathbf{Y} \mid \{\boldsymbol{\theta}_i, \boldsymbol{\delta}_i\}_{i=1}^{N_D}, \mathbf{Q}_{\boldsymbol{\delta}}, \mathbf{X})$ is the likelihood function of the full data set comprising all partitions of the data, which is expressed based on Eq. (4) as follows:

$$p(\mathbf{Y} \mid \{\boldsymbol{\theta}_i, \boldsymbol{\delta}_i\}_{i=1}^{N_D}, \mathbf{Q}_{\boldsymbol{\delta}}, \mathbf{X}) = N\left(\mathbf{Y} \mid \mathbf{f}(\mathbf{X}, \boldsymbol{\Theta}), \mathbf{K}(\boldsymbol{\Phi})\right) \tag{19}$$

where the vectors $\boldsymbol{\Theta} = [\boldsymbol{\theta}_1^T \ \cdots \ \boldsymbol{\theta}_{N_D}^T]^T$ and $\boldsymbol{\Phi} = [\boldsymbol{\phi}_1^T \ \cdots \ \boldsymbol{\phi}_{N_D}^T]^T$ are expanded vectors, containing the physical parameters and kernel parameters corresponding to all partitions, respectively. Note that the elements of the Gaussian distribution in Eq. (19) are given as

$$\mathbf{Y} = \begin{bmatrix} \mathbf{y}_1 \\ \mathbf{y}_2 \\ \mathbf{y}_3 \\ \vdots \\ \mathbf{y}_{N_D} \end{bmatrix} ; \mathbf{f}(\mathbf{X}, \boldsymbol{\Theta}) = \begin{bmatrix} f(\mathbf{x}_1, \boldsymbol{\theta}_1) \\ f(\mathbf{x}_2, \boldsymbol{\theta}_2) \\ f(\mathbf{x}_3, \boldsymbol{\theta}_3) \\ \vdots \\ f(\mathbf{x}_{N_D}, \boldsymbol{\theta}_{N_D}) \end{bmatrix} ; \mathbf{K}(\boldsymbol{\Phi}) = \begin{bmatrix} \mathbf{K}_1 & \mathbf{k}_{1,2} & \mathbf{k}_{1,3} & \cdots & \mathbf{k}_{1,N_D} \\ \mathbf{k}_{1,2}^T & \mathbf{K}_2 & \mathbf{k}_{2,3} & \cdots & \mathbf{k}_{2,N_D} \\ \mathbf{k}_{1,3}^T & \mathbf{k}_{2,3}^T & \mathbf{K}_3 & \cdots & \vdots \\ \vdots & \vdots & \vdots & \ddots & \mathbf{k}_{N_D-1,N_D} \\ \mathbf{k}_{1,N_D}^T & \mathbf{k}_{2,N_D}^T & \cdots & \mathbf{k}_{N_D-1,N_D}^T & \mathbf{K}_{N_D} \end{bmatrix} \tag{20}$$

In this equation, the elements of the covariance matrix is calculated based on the kernel function. This means that the elements $[\mathbf{K}_i]_{pq}$ and $[\mathbf{k}_{i,j}]_{pq}$ are calculated from the kernel function $k_\varepsilon(\boldsymbol{\zeta}_p, \boldsymbol{\zeta}'_q; \boldsymbol{\phi}_i)$, wherein the kernel parameters $\boldsymbol{\phi}_i$ are replaced when $i < j$.

Substituting the likelihood function from Eq. (19) and the prior distribution from Eq. (17) gives the joint posterior distribution:

$$p(\{\boldsymbol{\delta}_i, \boldsymbol{\theta}_i\}_{i=1}^{N_D}, \mathbf{Q}_{\boldsymbol{\delta}} \mid \mathbf{X}, \mathbf{Y}) \propto N(\mathbf{Y} \mid \mathbf{f}(\mathbf{X}, \boldsymbol{\Theta}), \mathbf{K}(\boldsymbol{\Phi})) \left[ \prod_{i=1}^{N_D} N(\boldsymbol{\theta}_i \mid \boldsymbol{\mu}_{\boldsymbol{\theta}_i}, \boldsymbol{\Sigma}_{\boldsymbol{\theta}_i}) N(\boldsymbol{\delta}_i \mid \boldsymbol{\delta}_{i-1}, \mathbf{Q}_{\boldsymbol{\delta}}) \right] p(\mathbf{Q}_{\boldsymbol{\delta}})$$

$$\propto N(\mathbf{Y} \mid \mathbf{f}(\mathbf{X}, \boldsymbol{\Theta}), \mathbf{K}(\boldsymbol{\Phi})) N(\boldsymbol{\Theta} \mid \boldsymbol{\mu}_{\boldsymbol{\Theta}}, \boldsymbol{\Sigma}_{\boldsymbol{\Theta}}) \left[ \prod_{i=1}^{N_D} N(\boldsymbol{\delta}_i \mid \boldsymbol{\delta}_{i-1}, \mathbf{Q}_{\boldsymbol{\delta}}) \right] p(\mathbf{Q}_{\boldsymbol{\delta}}) \quad (21)$$

where $\boldsymbol{\mu}_{\boldsymbol{\Theta}} = [\boldsymbol{\mu}_{\boldsymbol{\theta}_1}^T, ..., \boldsymbol{\mu}_{\boldsymbol{\theta}_{N_D}}^T]^T$ and $\boldsymbol{\Sigma}_{\boldsymbol{\Theta}} = \text{block-diag}(\boldsymbol{\Sigma}_{\boldsymbol{\theta}_1}, ..., \boldsymbol{\Sigma}_{\boldsymbol{\theta}_{N_D}})$ comprise the hyper-parameters of all data partitions. Like the single partition formulation in Section 2, the simultaneous estimation of $\boldsymbol{\theta}_i$ and $\{\boldsymbol{\mu}_{\boldsymbol{\theta}_i}, \boldsymbol{\Sigma}_{\boldsymbol{\theta}_i}\}$ can be cumbersome, so it is beneficial to marginalize $\boldsymbol{\theta}_i$'s from the posterior distribution giving:

$$p(\{\boldsymbol{\delta}_i\}_{i=1}^{N_D}, \mathbf{Q}_{\boldsymbol{\delta}} \mid \mathbf{X}, \mathbf{Y}) \propto p(\mathbf{Y} \mid \{\boldsymbol{\delta}_i\}_{i=1}^{N_D}, \mathbf{Q}_{\boldsymbol{\delta}}, \mathbf{X}) \left[ \prod_{i=1}^{N_D} N(\boldsymbol{\delta}_i \mid \boldsymbol{\delta}_{i-1}, \mathbf{Q}_{\boldsymbol{\delta}}) \right] p(\mathbf{Q}_{\boldsymbol{\delta}}) \quad (22)$$

where the marginalized likelihood function, $p(\mathbf{Y} \mid \{\boldsymbol{\delta}_i\}_{i=1}^{N_D}, \mathbf{Q}_{\boldsymbol{\delta}}, \mathbf{X})$, is given as

$$p(\mathbf{Y} \mid \{\boldsymbol{\delta}_i\}_{i=1}^{N_D}, \mathbf{Q}_{\boldsymbol{\delta}}, \mathbf{X}) = \int_{\boldsymbol{\Theta}} N(\mathbf{Y} \mid \mathbf{f}(\mathbf{X}, \boldsymbol{\Theta}), \mathbf{K}(\boldsymbol{\Phi})) N(\boldsymbol{\Theta} \mid \boldsymbol{\mu}_{\boldsymbol{\Theta}}, \boldsymbol{\Sigma}_{\boldsymbol{\Theta}}) d\boldsymbol{\Theta}$$

$$\approx N(\mathbf{Y} \mid \mathbf{f}(\mathbf{X}, \boldsymbol{\mu}_{\boldsymbol{\Theta}}), \mathbf{K}(\boldsymbol{\Phi}) + \mathbf{J} \boldsymbol{\Sigma}_{\boldsymbol{\Theta}} \mathbf{J}^T) \quad (23)$$

This integral is calculated using the results of Theorem 1 of Appendix (A), where the linearization with respect to $\boldsymbol{\Theta}$ around $\boldsymbol{\mu}_{\boldsymbol{\Theta}}$ is exercised. Moreover, it is noted that $\mathbf{J} = \text{block-diag}(\mathbf{J}_1, ..., \mathbf{J}_{N_D})$ consists of $\mathbf{J}_i$'s given as $\mathbf{J}_i = \partial f(\mathbf{x}_i, \boldsymbol{\theta}_i)/\partial \boldsymbol{\theta}_i^T \big|_{\boldsymbol{\theta}_i = \boldsymbol{\mu}_{\boldsymbol{\theta}_i}}$. On this basis, the marginal likelihood function can be expressed in an expanded format:

$$p(\mathbf{Y} \mid \{\boldsymbol{\delta}_i\}_{i=1}^{N_D}, \mathbf{Q}_{\boldsymbol{\delta}}, \mathbf{X})$$

$$= N\left( \begin{bmatrix} \mathbf{y}_1 \\ \mathbf{y}_2 \\ \mathbf{y}_3 \\ \vdots \\ \mathbf{y}_{N_D} \end{bmatrix} \left| \begin{bmatrix} f(\mathbf{x}_1, \boldsymbol{\mu}_{\boldsymbol{\theta}_1}) \\ f(\mathbf{x}_2, \boldsymbol{\mu}_{\boldsymbol{\theta}_2}) \\ f(\mathbf{x}_3, \boldsymbol{\mu}_{\boldsymbol{\theta}_3}) \\ \vdots \\ f(\mathbf{x}_{N_D}, \boldsymbol{\mu}_{\boldsymbol{\theta}_{N_D}}) \end{bmatrix}, \begin{bmatrix} \mathbf{K}_1 + \mathbf{J}_1 \boldsymbol{\Sigma}_{\boldsymbol{\theta}_1} \mathbf{J}_1^T & \mathbf{k}_{1,2} & \mathbf{k}_{1,3} & \cdots & \mathbf{k}_{1,N_D} \\ \mathbf{k}_{1,2}^T & \mathbf{K}_2 + \mathbf{J}_2 \boldsymbol{\Sigma}_{\boldsymbol{\theta}_2} \mathbf{J}_2^T & \mathbf{k}_{2,3} & \cdots & \mathbf{k}_{2,N_D} \\ \mathbf{k}_{1,3}^T & \mathbf{k}_{2,3}^T & \mathbf{K}_3 + \mathbf{J}_3 \boldsymbol{\Sigma}_{\boldsymbol{\theta}_3} \mathbf{J}_3^T & \cdots & \vdots \\ \vdots & \vdots & \vdots & \ddots & \mathbf{k}_{N_D-1, N_D} \\ \mathbf{k}_{1,N_D}^T & \mathbf{k}_{2,N_D}^T & \cdots & \mathbf{k}_{N_D-1,N_D}^T & \mathbf{K}_{N_D} + \mathbf{J}_{N_D} \boldsymbol{\Sigma}_{\boldsymbol{\theta}_{N_D}} \mathbf{J}_{N_D}^T \end{bmatrix} \right. \right) \quad (24)$$

In this equation, the processing of $N_D$ data partitions cannot be performed separately for each partition. Since the number of unknown parameters grows linearly with the number of data partitions, the computational cost becomes progressively more costly. This issue can be circumvented by approximating the conditional distribution by imposing a Markovian dependence between data partitions and thereof parameters so that sequential data processing becomes practicable. Such a simplification implies that the correlation of faraway partitions can be ignored. This simplification is justifiable in the context of the MMTE kernel since the correlation exponentially decays with the distance. Based on this assumption, the full data set likelihood can be simplified into

$$p\left(\mathbf{Y} \mid \{\boldsymbol{\delta}_i\}_{i=1}^{N_D}, \mathbf{Q}_{\boldsymbol{\delta}}, \mathbf{X}\right) = \left[\prod_{k=2}^{N_D} p\left(\mathbf{y}_k \mid \{\mathbf{y}_i\}_{i=1}^{k-1}, \{\boldsymbol{\delta}_i\}_{i=1}^{N_D}, \mathbf{X}\right)\right] \times p\left(\mathbf{y}_1 \mid \{\boldsymbol{\delta}_i\}_{i=1}^{N_D}, \mathbf{X}\right)$$

$$\approx \left[\prod_{k=2}^{N_D} p\left(\mathbf{y}_k \mid \mathbf{y}_{k-1}, \mathbf{x}_{k-1}, \mathbf{x}_k, \boldsymbol{\delta}_{k-1}, \boldsymbol{\delta}_k\right)\right] \times p\left(\mathbf{y}_1 \mid \mathbf{x}_1, \boldsymbol{\delta}_1\right) \quad (25)$$

$$\approx \left[\prod_{k=2}^{N_D} \frac{p\left(\mathbf{y}_{k-1}, \mathbf{y}_k \mid \mathbf{x}_{k-1}, \mathbf{x}_k, \boldsymbol{\delta}_{k-1}, \boldsymbol{\delta}_k\right)}{p\left(\mathbf{y}_{k-1} \mid \mathbf{x}_{k-1}, \boldsymbol{\delta}_{k-1}\right)}\right] \times p\left(\mathbf{y}_1 \mid \mathbf{x}_1, \boldsymbol{\delta}_1\right)$$

From Eq. (24), the probability distributions required in Eq. (25) can be calculated from the general properties of multivariate Gaussian distributions, which yield [36]:

$$p(\mathbf{y}_1 \mid \mathbf{x}_1, \boldsymbol{\delta}_1) = N\left(\mathbf{y}_1 \mid f(\mathbf{x}_1, \boldsymbol{\mu}_{\boldsymbol{\theta}_1}), \mathbf{K}_1 + \mathbf{J}_1 \boldsymbol{\Sigma}_{\boldsymbol{\theta}_1} \mathbf{J}_1^T\right) \quad (26)$$

$$p(\mathbf{y}_{k-1} \mid \mathbf{x}_{k-1}, \boldsymbol{\delta}_{k-1}) = N\left(\mathbf{y}_{k-1} \mid f(\mathbf{x}_{k-1}, \boldsymbol{\mu}_{\boldsymbol{\theta}_{k-1}}), \mathbf{K}_{k-1} + \mathbf{J}_{k-1} \boldsymbol{\Sigma}_{\boldsymbol{\theta}_{k-1}} \mathbf{J}_{k-1}^T\right) \quad (27)$$

$$p(\mathbf{y}_{k-1}, \mathbf{y}_k \mid \mathbf{x}_{k-1}, \mathbf{x}_k, \boldsymbol{\delta}_{k-1}, \boldsymbol{\delta}_k) = N\left(\begin{bmatrix} \mathbf{y}_{k-1} \\ \mathbf{y}_k \end{bmatrix} \middle| \begin{bmatrix} f(\mathbf{x}_{k-1}, \boldsymbol{\mu}_{\boldsymbol{\theta}_{k-1}}) \\ f(\mathbf{x}_k, \boldsymbol{\mu}_{\boldsymbol{\theta}_k}) \end{bmatrix}, \begin{bmatrix} \mathbf{K}_{k-1} + \mathbf{J}_{k-1} \boldsymbol{\Sigma}_{\boldsymbol{\theta}_{k-1}} \mathbf{J}_{k-1}^T & \mathbf{k}_{k-1,k} \\ \mathbf{k}_{k-1,k}^T & \mathbf{K}_k + \mathbf{J}_k \boldsymbol{\Sigma}_{\boldsymbol{\theta}_k} \mathbf{J}_k^T \end{bmatrix}\right) \quad (28)$$

Based on Theorem 2 of Appendix (C), the likelihood function can be transformed into a Markovian format, given by

$$p\left(\mathbf{Y} \mid \{\boldsymbol{\delta}_i\}_{i=1}^{N_D}, \mathbf{Q}_{\boldsymbol{\delta}}, \mathbf{X}\right) = \prod_{i=1}^{N_D} N\left(\mathbf{y}_i \mid \boldsymbol{\mu}_{i|i-1}, \boldsymbol{\Sigma}_{i|i-1}\right) \quad (29)$$

where $\boldsymbol{\mu}_{i|i-1} \in \mathbb{R}^{N_o}$ and $\boldsymbol{\Sigma}_{i|i-1} \in \mathbb{R}^{N_o \times N_o}$ are respectively the conditional mean vector and covariance matrix, given for $i \geq 2$, as follows:

$$\boldsymbol{\mu}_{i|i-1} = f(\mathbf{x}_i; \boldsymbol{\mu}_{\boldsymbol{\theta}_i}) + \mathbf{k}_{i-1,i}^T \left(\mathbf{K}_{i-1} + \mathbf{J}_{i-1} \boldsymbol{\Sigma}_{\boldsymbol{\theta}_{i-1}} \mathbf{J}_{i-1}^T\right)^{-1} \left(\mathbf{y}_{i-1} - f(\mathbf{x}_{i-1}; \boldsymbol{\mu}_{\boldsymbol{\theta}_{i-1}})\right) \quad (30)$$

$$\boldsymbol{\Sigma}_{i|i-1} = \left(\mathbf{K}_i + \mathbf{J}_i \boldsymbol{\Sigma}_{\boldsymbol{\theta}_i} \mathbf{J}_i^T\right) - \mathbf{k}_{i-1,i}^T \left(\mathbf{K}_{i-1} + \mathbf{J}_{i-1} \boldsymbol{\Sigma}_{\boldsymbol{\theta}_{i-1}} \mathbf{J}_{i-1}^T\right)^{-1} \mathbf{k}_{i-1,i} \quad (31)$$

Based on this simplified likelihood function, the posterior distribution can be rewritten as

$$p\left(\{\boldsymbol{\delta}_i\}_{i=1}^{N_D}, \mathbf{Q}_{\boldsymbol{\delta}} \mid \mathbf{X}, \mathbf{Y}\right) = \left[\prod_{i=1}^{N_D} N(\mathbf{y}_i \mid \boldsymbol{\mu}_{i|i-1}, \boldsymbol{\Sigma}_{i|i-1}) N(\boldsymbol{\delta}_i \mid \boldsymbol{\delta}_{i-1}, \mathbf{Q}_{\boldsymbol{\delta}})\right] p(\mathbf{Q}_{\boldsymbol{\delta}}) \quad (32)$$

It should be noted that the mean and covariance are given for $i = 1$ by $\boldsymbol{\mu}_{1|0} = f(\mathbf{x}_1, \boldsymbol{\mu}_{\boldsymbol{\theta}_1})$ and $\boldsymbol{\Sigma}_{1|0} = \mathbf{K}_1 + \mathbf{J}_1 \boldsymbol{\Sigma}_{\boldsymbol{\theta}_1} \mathbf{J}_1^T$, respectively. Here, the posterior distribution benefits from a Markovian structure since the parameters and data of each partition only depends on their preceding counterparts. This property will be shown to be advantageous for developing a sequential computational algorithm.

### 3.2. Graphical Representation

A graphical representation of the proposed probabilistic model is displayed in Fig. 2 using plate notation [37]. The rectangular plate represents each data partition and embeds partition-specific parameters. The parameters considered fixed across-partitions, i.e., $\mathbf{Q}_{\boldsymbol{\mu}_\theta}$, $\mathbf{Q}_{\boldsymbol{\Sigma}_\theta}$, and $\mathbf{Q}_{\boldsymbol{\phi}}$, are left out of this plate and shown by the purple circles. The white circles refer to the unknown parameters, and the gray circle indicates each partition of the data. The arrows represent the conditional dependence of the parameters, and the curved arrows indicate the sequential dependence of the parameters across different partitions. As can be seen, the Markovian structure of the data partitions and data-specific parameters, i.e., $\mathbf{y}_i$, $\boldsymbol{\mu}_{\boldsymbol{\theta}_i}$, $\boldsymbol{\Sigma}_{\boldsymbol{\theta}_i}$, and $\boldsymbol{\phi}_i$, depend on their counterparts of the $(i-1)^{th}$ partition shown by the curved arrows.

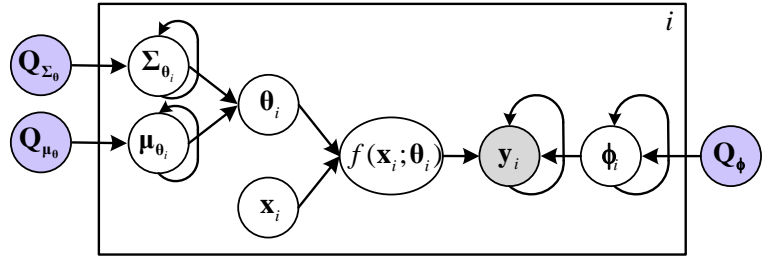

**Fig. 2.** Graphical representation of the proposed hierarchical probabilistic model

### 3.3. Posterior Predictive Distribution

Once the posterior distribution is calculated, it is possible to predict unobserved responses. This goal requires predicting statistical properties of the unobserved responses using the observed data. Let $\boldsymbol{\delta}_{N_D+1}$ denote the parameters of a new partition, which should be inferred from $N_D$ partitions. By virtue of the principle of total probability, the posterior predictive distribution of $\boldsymbol{\delta}_{N_D+1}$ can be calculated from

$$p(\boldsymbol{\delta}_{N_D+1} | \mathbf{X}, \mathbf{Y}) = \int_{\mathbf{Q}_{\boldsymbol{\delta}}} \int_{\boldsymbol{\delta}_{N_D}} N(\boldsymbol{\delta}_{N_D+1} | \boldsymbol{\delta}_{N_D}, \mathbf{Q}_{\boldsymbol{\delta}}) p(\boldsymbol{\delta}_{N_D}, \mathbf{Q}_{\boldsymbol{\delta}} | \mathbf{X}, \mathbf{Y}) d\boldsymbol{\delta}_{N_D} d\mathbf{Q}_{\boldsymbol{\delta}} \tag{33}$$

where the distribution $p(\boldsymbol{\delta}_{N_D}, \mathbf{Q}_{\boldsymbol{\delta}} | \mathbf{Y})$ should be calculated from Eq. (32) through a proper marginalization, which is to be discussed in Section 4. In this equation, the uncertainty of $\boldsymbol{\delta}_{N_D}$ and $\mathbf{Q}_{\boldsymbol{\delta}}$ is propagated to predict the unknown parameters $\boldsymbol{\delta}_{N_D+1}$. Subsequently, the unobserved response $\mathbf{y}_{N_D+1}$ can be calculated as

$$p(\mathbf{y}_{N_D+1} | \mathbf{x}_{N_D+1}, \mathbf{X}, \mathbf{Y}) = \int_{\boldsymbol{\delta}_{N_D+1}} p(\mathbf{y}_{N_D+1} | \mathbf{x}_{N_D+1}, \boldsymbol{\delta}_{N_D+1}) p(\boldsymbol{\delta}_{N_D+1} | \mathbf{X}, \mathbf{Y}) d\boldsymbol{\delta}_{N_D+1} \tag{34}$$

where $p(\mathbf{y}_{N_D+1} | \mathbf{x}_{N_D+1}, \boldsymbol{\delta}_{N_D+1})$ is the posterior predictive distribution given as $N\left(\mathbf{y}_{N_D+1} | \boldsymbol{\mu}_{N_D+1|N_D}, \boldsymbol{\Sigma}_{N_D+1|N_D}\right)$, where the mean and covariance are given by Eqs. (30-31). In these equations, the integrations over the multi-dimensional space of $\boldsymbol{\delta}_{N_D}$, $\boldsymbol{\delta}_{N_D+1}$, and $\mathbf{Q}_{\boldsymbol{\delta}}$ is computationally expensive, for which specific strategies will be provided in Section 4.

### 4. Computational Algorithm

In the context of Bayesian methods, the exact calculation of the posterior distribution is often intractable, and it is necessary to establish proper approximations. This problem is also the case with the posterior distribution obtained in Eq. (32), where a large number of parameters are involved, and they grow linearly with the number of partitions. Therefore, applying a sampling method can be a laborious approach owing to the complex dependence of the parameters, as indicated in Fig. 2.

In this paper, an optimization strategy is instead adopted to approximate the posterior distribution by ignoring the identification uncertainty of $\boldsymbol{\delta}_i$. Such a strategy can be justified for most practical cases, where the modeling errors are large such that the partition-to-partition variability is much larger than the identification uncertainty [38]. By doing so, the posterior distribution in Eq. (32) can be approximated as

$$p\left(\{\boldsymbol{\delta}_i\}_{i=1}^{N_D}, \mathbf{Q}_{\boldsymbol{\delta}} | \mathbf{X}, \mathbf{Y}\right) \approx \left[\prod_{i=1}^{N_D} \delta(\boldsymbol{\delta}_i - \hat{\boldsymbol{\delta}}_i) N(\boldsymbol{\delta}_i | \boldsymbol{\delta}_{i-1}, \mathbf{Q}_{\boldsymbol{\delta}})\right] p(\mathbf{Q}_{\boldsymbol{\delta}}) \tag{35}$$

where $\hat{\boldsymbol{\delta}}_i$ denotes the Most Probable Values (MPVs), and $\delta(\boldsymbol{\delta}_i - \hat{\boldsymbol{\delta}}_i)$ is Dirac delta function centered at $\hat{\boldsymbol{\delta}}_i$. The Markovian structure of the probabilistic model gives the opportunity to find the MPVs of $\boldsymbol{\delta}_i$'s sequentially by minimizing the negative log-likelihood function, $L(\boldsymbol{\delta}_i) = -\ln N(\mathbf{y}_i | \hat{\boldsymbol{\mu}}_{i|i-1}, \hat{\boldsymbol{\Sigma}}_{i|i-1})$, for each $\boldsymbol{\delta}_i$, given by

$$L(\boldsymbol{\delta}_i) = \frac{1}{2}\ln\left|\hat{\boldsymbol{\Sigma}}_{i|i-1}\right| + \frac{1}{2}(\mathbf{y}_i - \hat{\boldsymbol{\mu}}_{i|i-1})^T \hat{\boldsymbol{\Sigma}}_{i|i-1}^{-1}(\mathbf{y}_i - \hat{\boldsymbol{\mu}}_{i|i-1}) + cte. \tag{36}$$

where $|.|$ denotes the determinant of a matrix; $cte.$ stands for constant terms; the mean $\hat{\boldsymbol{\mu}}_{i|i-1}$ and the covariance matrix $\hat{\boldsymbol{\Sigma}}_{i|i-1}$ are obtained from Eqs. (30-31) while the parameters $\boldsymbol{\delta}_{i-1}$ is set to its MPVs $\hat{\boldsymbol{\delta}}_{i-1}$, obtained at the preceding step of the analysis.

Once the MPVs ($\hat{\boldsymbol{\delta}}_i$'s) are acquired for $i = \{1, \ldots, N_D\}$, the posterior distribution of the hyper-parameters $\mathbf{Q}_\delta$ can calculated from Eq. (35), giving:

$$p(\mathbf{Q}_\delta | \mathbf{Y}) \propto \left[\prod_{i=1}^{N_D} N(\hat{\boldsymbol{\delta}}_i | \hat{\boldsymbol{\delta}}_{i-1}, \mathbf{Q}_\delta)\right] p(\mathbf{Q}_\delta) \tag{37}$$

Note that this marginal posterior distribution is obtained by neglecting the identification uncertainty of $\boldsymbol{\delta}_i$. Considering $p(\mathbf{Q}_\delta)$ to be uniform, from this equation, the MPVs of $\mathbf{Q}_\delta$ can be calculated by maximizing Eq. (37) for $\mathbf{Q}_\delta$ or minimizing its negative-logarithm given by

$$L(\mathbf{Q}_\delta) = -\ln p(\mathbf{Q}_\delta | \mathbf{Y}) = \frac{N_D}{2}\ln|\mathbf{Q}_\delta| + \frac{1}{2}\left[\sum_{i=1}^{N_D}(\hat{\boldsymbol{\delta}}_i - \hat{\boldsymbol{\delta}}_{i-1})^T \mathbf{Q}_\delta^{-1}(\hat{\boldsymbol{\delta}}_i - \hat{\boldsymbol{\delta}}_{i-1})\right] + cte. \tag{38}$$

where $L(\mathbf{Q}_\delta)$ is calculated from Eq. (37). Taking derivatives of this expression with respect to $\mathbf{Q}_\delta$ allows writing:

$$\frac{\partial L(\mathbf{Q}_\delta)}{\partial \mathbf{Q}_\delta} = \frac{N_D}{2}\mathbf{Q}_\delta^{-1} - \frac{1}{2}\mathbf{Q}_\delta^{-1}\left[\sum_{i=1}^{N_D}(\hat{\boldsymbol{\delta}}_i - \hat{\boldsymbol{\delta}}_{i-1})(\hat{\boldsymbol{\delta}}_i - \hat{\boldsymbol{\delta}}_{i-1})^T\right]\mathbf{Q}_\delta^{-1} \tag{39}$$

Setting this equation equal to zero provides:

$$\hat{\mathbf{Q}}_\delta = \frac{1}{N_D}\sum_{i=1}^{N_D}(\hat{\boldsymbol{\delta}}_i - \hat{\boldsymbol{\delta}}_{i-1})(\hat{\boldsymbol{\delta}}_i - \hat{\boldsymbol{\delta}}_{i-1})^T \tag{40}$$

where $\hat{\mathbf{Q}}_\delta$ is the MPVs of the covariance matrix $\mathbf{Q}_\delta$.

Similar to the approximation introduced in Eq. (37), the posterior predictive distribution of Eq. (33) can be written as

$$p(\boldsymbol{\delta}_{N_D+1} | \mathbf{X}, \mathbf{Y}) \approx \int_{\mathbf{Q}_\delta}\int_{\boldsymbol{\delta}_{N_D}} N(\boldsymbol{\delta}_{N_D+1} | \boldsymbol{\delta}_{N_D}, \mathbf{Q}_\delta)\delta(\boldsymbol{\delta}_{N_D} - \hat{\boldsymbol{\delta}}_{N_D})\delta(\mathbf{Q}_\delta - \hat{\mathbf{Q}}_\delta)d\boldsymbol{\delta}_{N_D}d\mathbf{Q}_\delta \approx N(\boldsymbol{\delta}_{N_D+1} | \hat{\boldsymbol{\delta}}_{N_D}, \hat{\mathbf{Q}}_\delta) \tag{41}$$

Since this distribution is Gaussian, it is straightforward to sample it for predicting the response subjected to future input. For this purpose, from Eq. (34), we can write:

$$p(\mathbf{y}_{N_D+1} | \mathbf{x}_{N_D+1}, \mathbf{X}, \mathbf{Y}) = \int_{\boldsymbol{\delta}_{N_D+1}} N\left(\mathbf{y}_{N_D+1} | \boldsymbol{\mu}_{N_D+1|N_D}, \boldsymbol{\Sigma}_{N_D+1|N_D}\right) N(\boldsymbol{\delta}_{N_D+1} | \hat{\boldsymbol{\delta}}_{N_D}, \hat{\mathbf{Q}}_\delta) d\boldsymbol{\delta}_{N_D+1}$$
$$\approx \frac{1}{N_s}\sum_{m=1}^{N_s} N\left(\mathbf{y}_{N_D+1} | \boldsymbol{\mu}_{N_D+1|N_D}^{(m)}, \boldsymbol{\Sigma}_{N_D+1|N_D}^{(m)}\right) \tag{42}$$

where $\boldsymbol{\mu}_{N_D+1|N_D}^{(m)}$ and $\boldsymbol{\Sigma}_{N_D+1|N_D}^{(m)}$ are respectively the samples of the mean vector and covariance matrix, wherein $\boldsymbol{\delta}_{N_D}$ is replaced with its MPVs whereas $\boldsymbol{\delta}_{N_D+1}$ is replaced by its samples $\boldsymbol{\delta}_{N_D+1}^{(m)}$ drawn from $N(\boldsymbol{\delta}_{N_D+1} | \hat{\boldsymbol{\delta}}_{N_D}, \hat{\mathbf{Q}}_\delta)$. This Gaussian mixture distribution is known to have the following statistical moments [31]:

$$\mathbb{E}\left[\mathbf{y}_{N_D+1}\right] = \frac{1}{N_s}\sum_{m=1}^{N_s}\boldsymbol{\mu}_{N_D+1|N_D}^{(m)} \tag{43}$$

$$\mathbb{COV}\left[\mathbf{y}_{N_D+1}\right] = \frac{1}{N_s}\left[\sum_{m=1}^{N_s}\left(\boldsymbol{\mu}_{N_D+1|N_D}^{(m)}\boldsymbol{\mu}_{N_D+1|N_D}^{(m),T} + \boldsymbol{\Sigma}_{N_D+1|N_D}^{(m)}\right)\right] - \frac{1}{N_s^2}\left[\sum_{m=1}^{N_s}\boldsymbol{\mu}_{N_D+1|N_D}^{(m)}\right]\left[\sum_{m=1}^{N_s}\boldsymbol{\mu}_{N_D+1|N_D}^{(m)}\right]^T \quad (44)$$

where $\mathbb{E}\left[\mathbf{y}_{N_D+1}\right]$ and $\mathbb{COV}\left[\mathbf{y}_{N_D+1}\right]$ are the mean and covariance of the unobserved responses $\mathbf{y}_{N_D+1}$, respectively.

Algorithm 1 gives a systematic procedure to address the computation of the proposed probabilistic model. This procedure starts with progressively finding the MPVs of the partition-specific parameters for each data partition. Then, it continues with calculating the MPVs of $\mathbf{Q}_\delta$ and predicting unmeasured structural responses. In this algorithm, calculating the determinant and the matrix multiplication $\hat{\boldsymbol{\Sigma}}_{i|i-1}^{-1}(\mathbf{y}_i - \hat{\boldsymbol{\mu}}_{i|i-1})$ is a critical step required between the lines 4 and 5. Since the procedure must be repeated at each step of the maximization, it governs a substantial amount of the computation. This problem, referred to as the "Scalability" of GPs, implies that the computational cost grow cubically with the dimension of the full covariance matrix, i.e., $O((nN_o N_D)^3)$ [39]. However, the data partitioning exercised in the present study mitigates the computational burden up to a large extent since long-length time-history data can be processed separately for short segments. This implementation essentially implies that the computational cost can be reduced from $O((nN_o N_D)^3)$ to $O(N_D(nN_o)^3)$, offering significant computational savings.

Additionally, calculating the determinant and inverse of $\hat{\boldsymbol{\Sigma}}_{i|i-1}$ might encounter ill-posedness as the number of data points within each partition $nN_o$ increases. To circumvent this problem, the Singular Value Decomposition (SVD) of $\hat{\boldsymbol{\Sigma}}_{i|i-1}$ is used for truncating redundant dimensions and reducing the dimensionality of the covariance matrix. In order to specify an appropriate level of truncation, the slope of the singular values decay can be checked against a predefined tolerance of 0.1-1% of the largest singular value.

**Algorithm 1.**
Calculation of the MPVs of the unknown parameters and predictions of unmeasured responses

1: Set the initial values of $\hat{\boldsymbol{\delta}}_0$, comprising $\hat{\boldsymbol{\mu}}_{\boldsymbol{\theta}_0}$, $\hat{\boldsymbol{\Sigma}}_{\boldsymbol{\theta}_0}$, and $\hat{\boldsymbol{\phi}}_0$.

2: **For** $i = 1 : N_D$ (Data Partitions) **{**

3: Find the MPVs $\hat{\boldsymbol{\delta}}_i = \{\hat{\boldsymbol{\mu}}_{\boldsymbol{\theta}_i}, \hat{\boldsymbol{\Sigma}}_{\boldsymbol{\theta}_i}, \hat{\boldsymbol{\phi}}_i\}$ by minimizing the log-likelihood function $L(\boldsymbol{\delta}_i)$ given by Eq. (36)

4: **} End For**

5: Approximate the MPVs of $\mathbf{Q}_\delta$ by the expression $\hat{\mathbf{Q}}_\delta \approx \frac{1}{N_D}\sum_{i=1}^{N_D}(\hat{\boldsymbol{\delta}}_i - \hat{\boldsymbol{\delta}}_{i-1})(\hat{\boldsymbol{\delta}}_i - \hat{\boldsymbol{\delta}}_{i-1})^T$.

6: Draw samples $\boldsymbol{\delta}_{N_D+1}^{(m)}$ from the distribution $N(\boldsymbol{\delta}_{N_D+1} | \hat{\boldsymbol{\delta}}_{N_D}, \hat{\mathbf{Q}}_\delta)$.

7: Compute the mean and covariance of unobserved responses using Eqs. (43-44).

## 5. Illustrative Examples
### 5.1. IASE-ASCE Benchmark Problem

The SHM benchmark structure is used to demonstrate the proposed methodology. The dimensions and mechanical properties of this structure are available in the literature [40]. As shown in Fig. 3(a), the structure is a steel-braced frame, constructed in four stories and covered by concrete slabs. It is subjected to a concentrated force acting on the 4[th] floor along the $x$ direction. The dynamical force $P_x(t)$ is considered GWN process with zero mean and 10kN standard deviation. For this structure, the out-of-plane deformations of the slabs are significantly smaller than the in-plane displacements and rotations. Additionally, the axial elongation of the columns is negligible compared to bending and shear deformations. Therefore, it is admissible to neglect the axial displacement of columns and simulate the structure by the 12-DOF lumped-mass column shown in Fig. 3(b). This simplification implies that each floor can have translational displacements along both $x$- and $y$-axes, as well as a rotational deformation along the $z$-axis. Based on this structural model, the elements of the mass and stiffness matrices ($\mathbf{M}$ and $\mathbf{K}$) were calculated. The damping matrix is considered proportional to the mass and stiffness matrices, given by $\mathbf{C} = \alpha \mathbf{K} + \beta \mathbf{M}$, where $\alpha = 0.02$ and $\beta = 2 \times 10^{-5}$. This damping matrix creates modal damping ratios of order 0.1-5%.

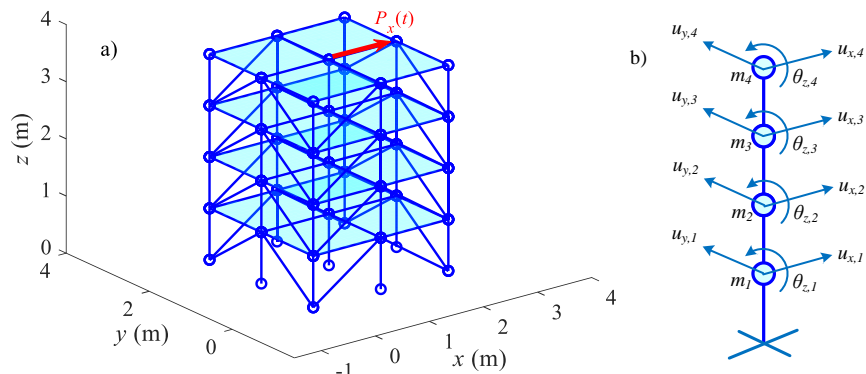

**Fig. 3.** (a) A 3D view of the IASE-ASCE benchmark structure (b) 12-DOF lumped mass model used for inference

Having made the above assumptions, the synthetic response of the structure is generated and recorded for an adequately long period using $\Delta t = 0.001 s$ sampling interval. The acceleration responses of the 2$^{nd}$ and 3$^{rd}$ floors ($\ddot{u}_{x,2}$ and $\ddot{u}_{x,3}$) are considered as measured quantities, contaminated with zero-mean additive GWN with a standard deviation equal to 5% Root-Mean-Square (RMS) of the noise-free signals. This simulated data is next partitioned into $N_D$ segments and used for model inference and response predictions.

The physics-based model used for simulating the dynamical responses is assumed to have the same mass matrix as the original structure. The stiffness matrix is partially unknown, whose elements $\theta_1 k_{x,2}$ and $\theta_2 k_{x,3}$ are to be determined. This means that the physical parameters $\boldsymbol{\theta} = [\theta_1 \; \theta_2]^T$ should be identified from the data. However, the damping matrix is considered to deviate from that of the structure using $\alpha' = 0.03$ and $\beta' = \beta = 2 \times 10^{-5}$. Moreover, the physical parameters are considered variable across data partitions, drawn from $\boldsymbol{\theta}_i \underset{iid}{\sim} N(\mathbf{0}, 0.01\mathbf{I})$, where $i = \{1, 2, ..., N_D\}$. This misspecification aims to simulate systematic errors when describing the observed dynamical responses.

The accuracy and complexity of the MMTE kernel function is governed by the number of incorporated cosine functions ($m$). In Fig. 4(a), the PSD of the prediction errors is indicated, wherein three dominant modes at frequencies 9, 40, and 48Hz can be observed. This initially suggests that $m = 3$ can be a good choice. To investigate this issue further, Bayesian Information Criteria (BIC) is calculated when $m = \{1, 2, 3, 4\}$ modes are incorporated into the kernel function. As shown in Fig. 4(b), the maximum BIC score is obtained for $m = 3$ when the three modes 9, 40, and 48Hz are used. Thus, the selection of these three modes can be supported intuitively and theoretically.

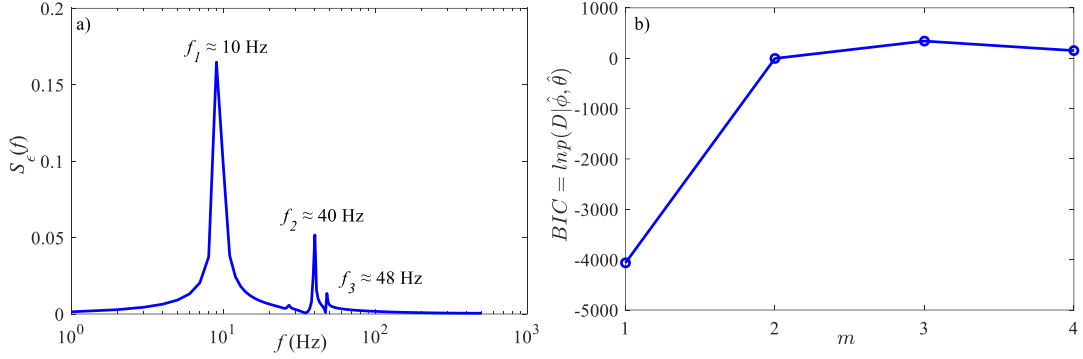

**Fig. 4.** (a) PSD of the discrepancy between the model and actual acceleration responses of the 3$^{rd}$ floor; (b) BIC score compared to the number of kernel modes

Having considered the above settings, Algorithm 1 is used to identify the unknown parameters of the model and kernel machines. The estimated mean of the kernel parameters are presented in Table 1 using different choices of the number of data points ($n$) and partitions ($N_D$). Estimations of $\omega_k/(2\pi)$, where $k=\{1,2,3\}$, are close to those observed in Fig. 4(a), and they are almost invariant regardless of the number of data points or partitions. Conversely, estimations of $\sigma_k^2$, $\sigma_n^2$, and $1/\ell_k^2$ vary depending on the data length $n$ and the number of partitions $N_D$. This observation is natural in the sense that the length of data (or the number of data points) govern temporal characteristics of the kernel function.

**Table 1.**
Estimated mean of the unknown parameters obtained from posterior predictive distribution

| $nN_D$ | $n$ | $N_D$ | $\sigma_1^2$ | $1/\ell_1^2$ | $\omega_1/(2\pi)$ | $\sigma_2^2$ | $1/\ell_2^2$ | $\omega_2/(2\pi)$ | $\sigma_3^2$ | $1/\ell_3^2$ | $\omega_3/(2\pi)$ | $\sigma_n^2$ |
|---|---|---|---|---|---|---|---|---|---|---|---|---|
| 3000 | 500 | 6 | 0.88 | 3.95 | 9.84 | 0.99 | 2.57 | 39.39 | 0.87 | 2.98 | 47.79 | 3×10⁻⁶ |
| | 1000 | 3 | 0.07 | 0.50 | 10.66 | 0.89 | 0.48 | 39.82 | 0.86 | 2.72 | 47.79 | 4×10⁻⁶ |
| 9000 | 300 | 30 | 1.18 | 8.45 | 10.25 | 1.18 | 9.50 | 39.40 | 1.16 | 7.07 | 46.18 | 2×10⁻⁵ |
| | 1000 | 9 | 0.45 | 1.24 | 10.83 | 0.99 | 1.03 | 39.80 | 1.04 | 1.43 | 47.90 | 7×10⁻⁶ |

Table 2 presents the mean and standard deviation of the stiffness parameters identified for different choices of the number of data points and partitions. The estimated mean values are very close to the actual values ($\mu_{\theta_1} = \mu_{\theta_2} = 1$), and the estimated standard deviations closely agree with the actual values ($\sigma_{\theta_1} = \sigma_{\theta_2} = 0.01$), confirming the validity of the results. Additionally, the mean and standard deviation of the stiffness parameters remain consistent regardless of the number of data points and partitions. This asymptotic behavior is desirable as it resolves a significant shortcoming of the classical Bayesian approach [1], where the posterior uncertainty reduces with the number of data points and partitions.

**Table 2.**
Estimations of the mean and standard deviation of the stiffness parameters obtained based on the posterior predictive distribution

| $nN_D$ | $n$ | $N_D$ | $\mu_{\theta_1}$ | $\mu_{\theta_2}$ | $\sigma_{\theta_1}$ | $\sigma_{\theta_2}$ |
|---|---|---|---|---|---|---|
| 3000 | 500 | 6 | 0.998 | 1.020 | 0.0101 | 0.0099 |
| | 1000 | 3 | 1.002 | 0.997 | 0.0098 | 0.0082 |
| 9000 | 300 | 30 | 1.008 | 1.000 | 0.0103 | 0.0102 |
| | 1000 | 9 | 1.006 | 1.004 | 0.0092 | 0.0102 |

Having identified the unknown parameters, we propagate uncertainties for predicting dynamical responses based on Algorithm 1. Predictions of the acceleration response of the 4$^{th}$ DOF ($\ddot{u}_{x,2}(t)$) are shown in Fig. 5. The first 10s segment of the measured response is regarded as the training data set, and the next 10s data is treated as the held-out data set. For better visualization, four zoomed windows are provided. As can be seen, the estimated mean response agrees with the measured response, and the uncertainty bounds indicated by the shaded areas reasonably account for potential errors. A similar pattern can be observed in Fig. 6 for the 7$^{th}$ DOF ($\ddot{u}_{x,3}(t)$), emphasizing the accuracy of the proposed method.

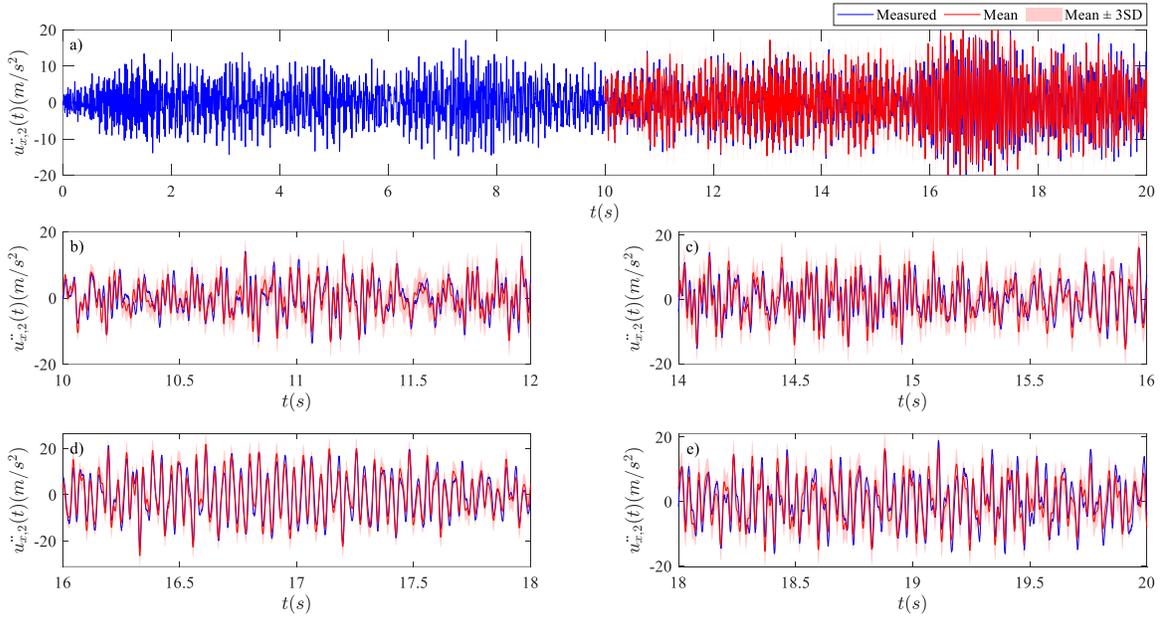

**Fig. 5.** Predictions of the acceleration response of the 2$^{nd}$ floor along the *x*-axis when using 10s data partitioned into $N_D = 10$ segments, each comprising $n = 1000$ time samples

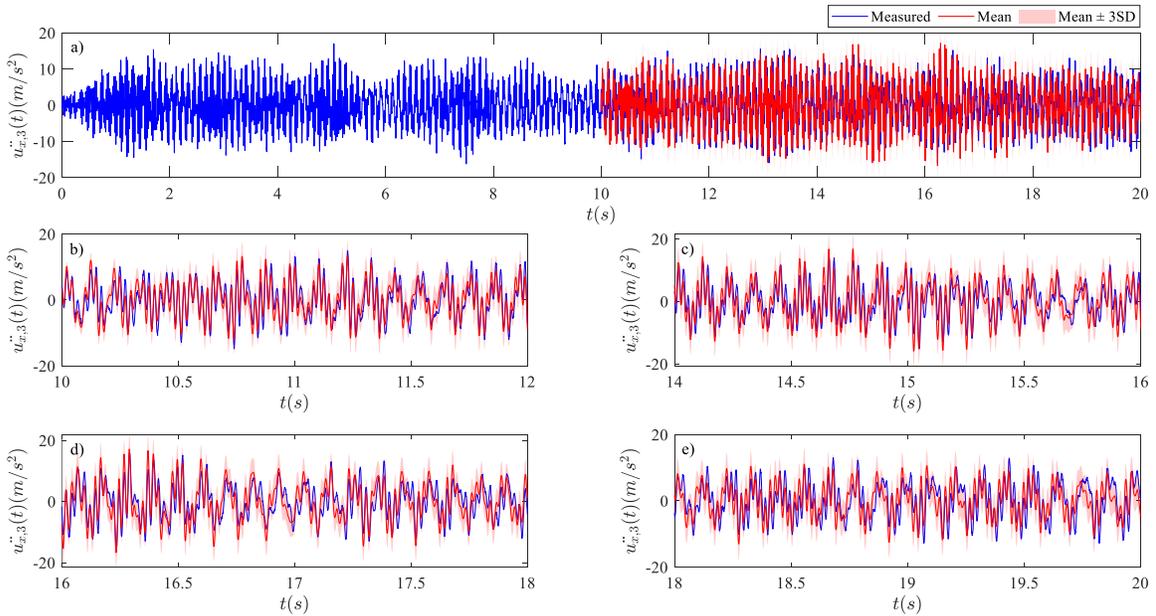

**Fig. 6.** Predictions of the acceleration response of the 3$^{rd}$ floor along the *x*-axis when using 10s data partitioned into $N_D = 10$ segments, each comprising $n = 1000$ time samples

Fig. 7 compares predictions of $\ddot{u}_{x,2}(t)$ based on the classical Bayesian approach (e.g., [1]) and the proposed method. Fig. 7(a) shows that the classical framework does not account for potential discrepancies between the estimated mean and the measured responses as the uncertainty bounds appear thinner than required. In contrast, Fig. 7(b) shows that the estimated mean and the uncertainty bounds (±3SD) calculated through the proposed framework well account for the errors caused by model misspecification.

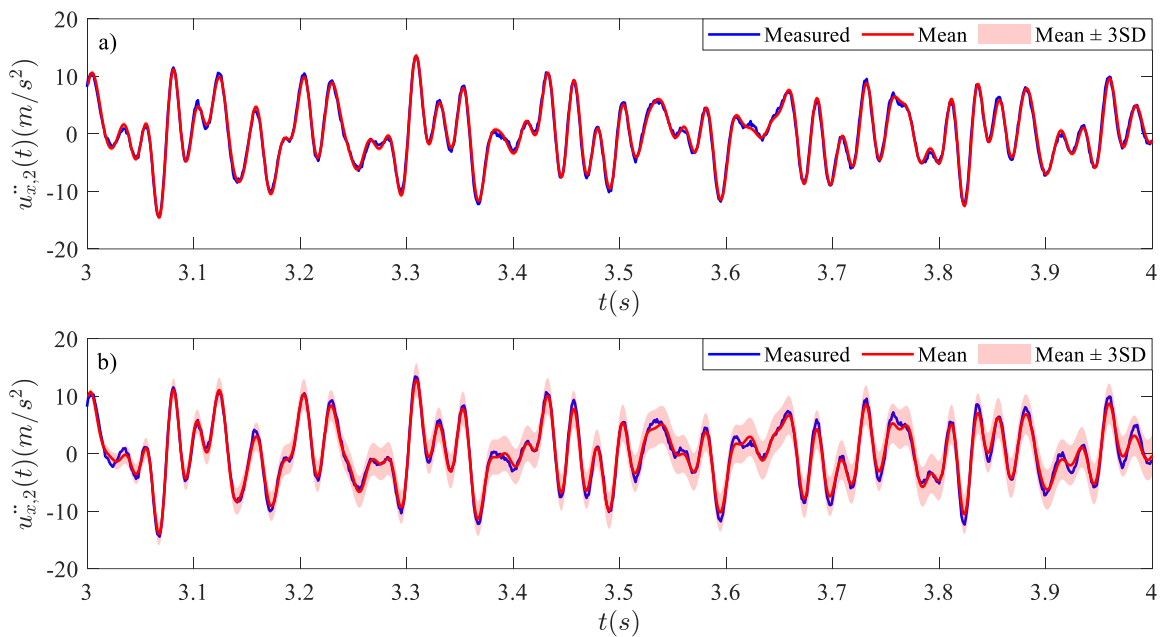

**Fig. 7.** Predictions of the acceleration time-history response of the 2$^{nd}$ floor along the *x*-axis using (a) Classical Bayesian framework (b) proposed framework

### 5.2. Experimental Example

In this section, we test the proposed framework using experimental data. For this purpose, a shaking-table test is performed to excite the base of a small-scale three-story structure shown in Fig. 8(a). The base and the three stories acceleration responses along the *x*-axis are measured subjected to a broadband GWN acceleration along the *x*-axis. A three-story shear frame shown in Fig. 8(b) can model the dynamical behavior of the structure. The applied excitation is plotted in Fig. 9(a), and the measured acceleration response of the 3$^{rd}$ floor is showcased in Fig. 9(b). The time-history acceleration responses of the three floors are recorded at 0.005s time intervals.

The mass of the structural model shown in Fig. 9(b) is considered lumped at each floor, originating from the thick floor plates. The mass of the columns and beams is relatively negligible. Thus, the mass matrix is diagonal having elements $m_1 = 5.63$ kg, $m_2 = 6.03$ kg, and $m_3 = 4.66$ kg. The damping matrix ($\hat{\mathbf{C}}$) is considered to be proportional to the mass matrix, given by

$$\hat{\mathbf{C}} = \sum_{i=1}^{3}\left(2\hat{\varpi}_i \hat{\zeta}_i \frac{\hat{\mathbf{M}}\hat{\boldsymbol{\phi}}_i \hat{\boldsymbol{\phi}}_i^T \hat{\mathbf{M}}}{\hat{\boldsymbol{\phi}}_i^T \hat{\mathbf{M}}\hat{\boldsymbol{\phi}}_i}\right) \tag{45}$$

where the modal damping ratios are obtained from [41,42], considered $\hat{\zeta}_1 = 2.39\%$, $\hat{\zeta}_2 = 0.87\%$, and $\hat{\zeta}_3 = 0.65\%$; $\hat{\mathbf{M}} = diag(m_1, m_2, m_3)$ is the mass matrix; $\hat{\varpi}_1 = 114.41$ rad/s, $\hat{\varpi}_2 = 81.82$ rad/s, and $\hat{\varpi}_3 = 28.56$ rad/s are the nominal modal frequencies; $\hat{\boldsymbol{\phi}}_i \in \mathbb{R}^3$ is the $i^{th}$ nominal mode shape corresponding to modal frequency $\hat{\varpi}_i$. Unlike the known matrices $\hat{\mathbf{M}}$ and $\hat{\mathbf{C}}$, the stiffness matrix is considered to be unknown, described as

$$\mathbf{K}(\boldsymbol{\theta}) = \begin{bmatrix} \theta_1 + \theta_2 & -\theta_2 & 0 \\ -\theta_2 & \theta_2 + \theta_3 & -\theta_3 \\ 0 & -\theta_3 & \theta_3 \end{bmatrix} \tag{46}$$

where $\boldsymbol{\theta} = \begin{bmatrix} \theta_1 & \theta_2 & \theta_3 \end{bmatrix}^T$ is a parameter vector, comprising the stiffness of each floor. The primary objective is to identify the unknown parameters using the acceleration responses of the 1$^{st}$ and the 3$^{rd}$ floors using the proposed methodology, and the secondary objective is to predict unobserved responses.

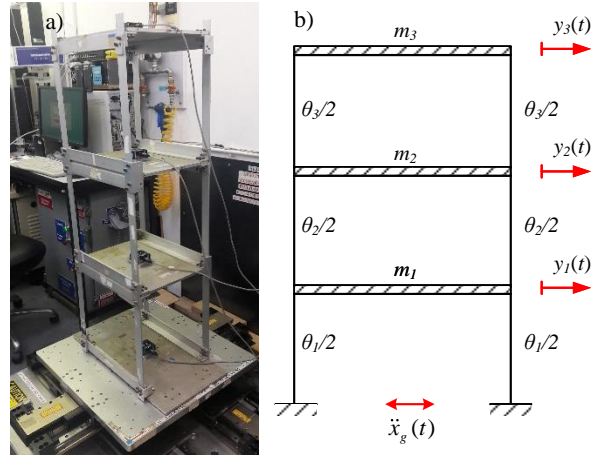

**Fig. 8.** (a) Three-story frame tested on a shaking-table (b) 3-DOF linear structural model

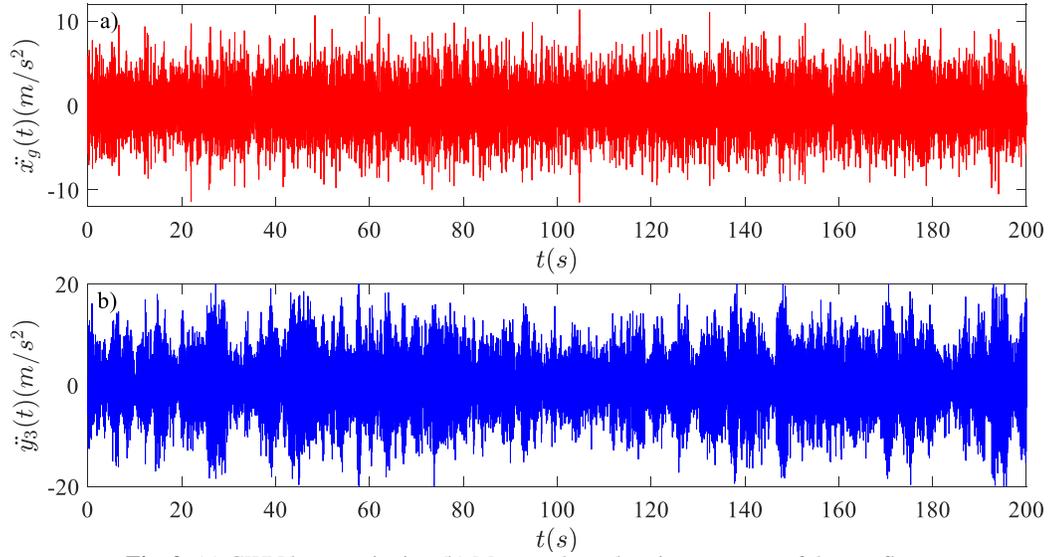

**Fig. 9.** (a) GWN base excitation (b) Measured acceleration response of the top floor

Based on Algorithm 1, TMCMC sampling method [43–45] is used to draw $N_s = 10000$ samples of $p(\boldsymbol{\delta}_{N_D+1} | D_{N_D})$, considering $n = 500$ and $N_D = 20$. Fig. 10 visualizes the posterior predictive distribution $p(\boldsymbol{\mu}_{\boldsymbol{\theta}, N_D+1} | D_{N_D})$ based on the Markovian samples. The dashed red lines show the reference values reported in [41], which appear close to the mode of the calculated distribution. Moreover, the estimated stiffness of the first floor is calculated considerably smaller than those of the second and third floors. This behavior is suspected to be due to a loose connection of the first stories' column to the base plate compared to other floors.

Fig. 11 shows the posterior predictive distribution of the kernel function parameters, $p(\boldsymbol{\phi}_{N_D+1} | D_{N_D})$. The kernel frequencies can be approximated as $\omega_1 \approx 115$ rad/s, $\omega_2 \approx 82$ rad/s, and $\omega_3 \approx 28$ rad/s. These values one-to-one correspond to the modal frequencies of the structure mentioned earlier as $\hat{\varpi}_1$, $\hat{\varpi}_2$, and $\hat{\varpi}_3$. The correlation lengths fall within the interval $\ell_i \in [1.0 - 2.0], \forall i = \{1, 2, 3\}$, indicating fast decay of temporal correlation over time.

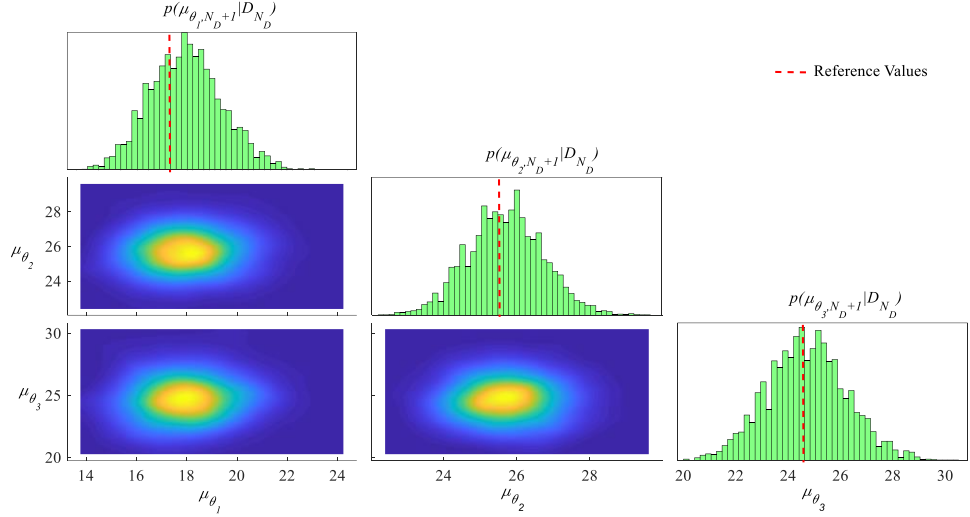

**Fig. 10**. Posterior predictive distribution $p(\boldsymbol{\mu}_{\theta,N_D+1} | D_{N_D})$ computed based on 20 sets of vibration data

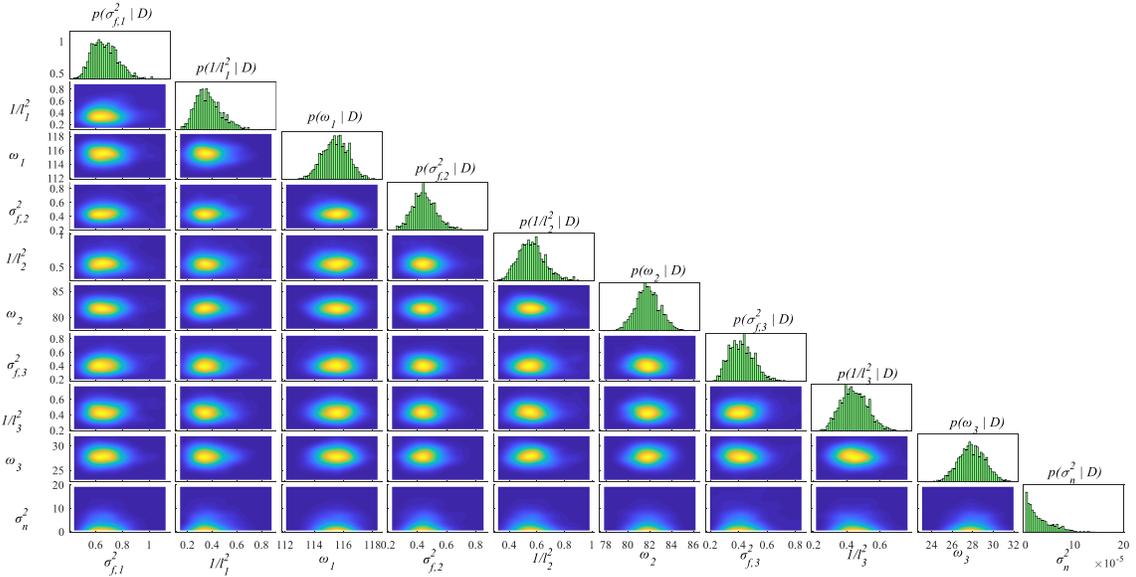

**Fig. 11**. Posterior predictive distribution of the kernel parameters computed based on 20 sets of vibration data

Table 3 shows the posterior mean and standard deviation of the structural parameters, identified considering different number of data points ($n$) and partitions ($N_D$). It can be seen that the stiffness of the first floor is identified smaller than the other two floors, which can originate from the loosening the bolts at the base. The estimated mean values also appear to be close to those reported in previous works [41]. Additionally, it is notable that, regardless of the values of $n$ and $N_D$, the estimated mean and standard deviations converge almost to the same values. This asymptotic behavior is desirable as the posterior distributions are insensitive to the number of data points and the redundancy of the measured information.

The convergence trend of the statistical properties of the structural parameters can be seen in Fig. 12 for $n = 4000$. The mean ($\hat{\mu}_{\theta_i}$) and standard deviation ($\hat{\sigma}_{\theta_i}$) estimated from each data partition is indicated by the error bars. The posterior predictive distribution $p(\boldsymbol{\delta}_{i+1} | D_i)$ is calculated for $i = \{1, 2, ..., 20\}$, giving a sense how the parameters vary as the information of different partitions accumulates. Based on this distribution, the estimated mean and standard deviation of the stiffness parameters are displayed in Fig. 12 through the red line and the shaded area, respectively. As shown, the uncertainty bounds can account for the variability of estimated mean values and standard deviations across different partitions of data.

**Table 3.**
Posterior mean and standard deviation of the stiffness parameters calculated for varying choices of the number of data points or partitions

| $nN_D$ | $n$ | $N_D$ | $\hat{\mu}_{\theta_1}$ | $\hat{\sigma}_{\theta_1}$ | $\hat{\mu}_{\theta_2}$ | $\hat{\sigma}_{\theta_2}$ | $\hat{\mu}_{\theta_3}$ | $\hat{\sigma}_{\theta_3}$ |
|---|---|---|---|---|---|---|---|---|
| 10000 | 1000 | 10 | 17.98 | 0.0867 | 25.58 | 0.0804 | 24.97 | 0.0938 |
|  | 500 | 20 | 17.95 | 0.0945 | 25.55 | 0.1011 | 24.96 | 0.1060 |
| 25000 | 1000 | 25 | 17.99 | 0.0975 | 25.54 | 0.1041 | 24.99 | 0.1105 |
|  | 500 | 50 | 17.96 | 0.1069 | 25.54 | 0.0896 | 24.96 | 0.1075 |

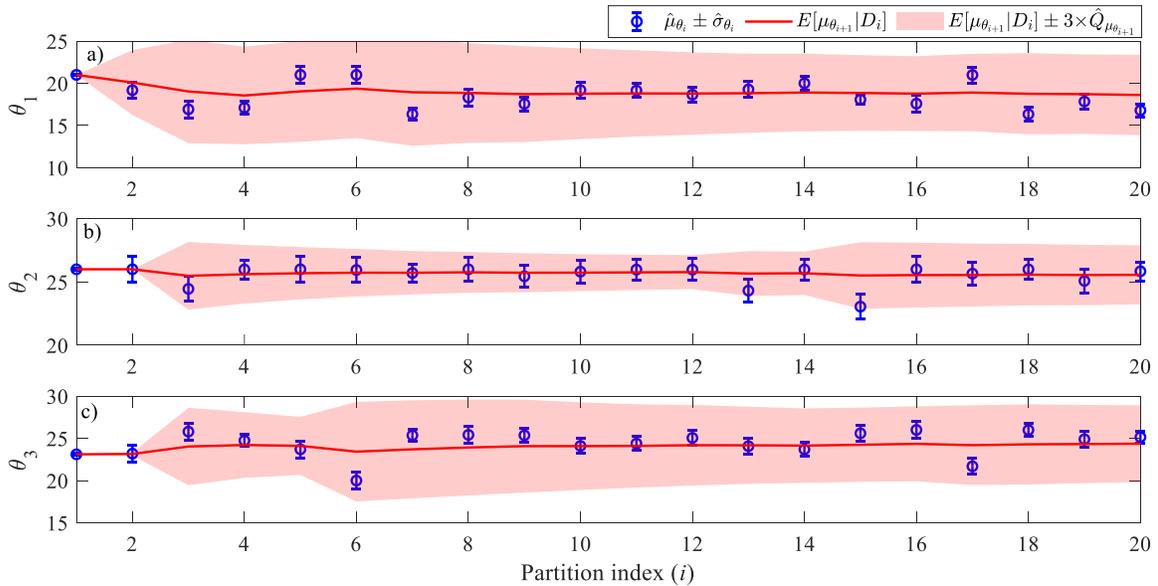

**Fig. 12**. Variation of the posterior distribution of the stiffness parameters with the number of partitions

Predictions of the acceleration responses of the 1st and 3rd floors are plotted in Fig. 13. As can be seen, the estimated mean response closely follows the measured signal. Moreover, the shaded area representing 99% confidence interval (±3SD) fairly contains potential discrepancy between the measured and mean responses.

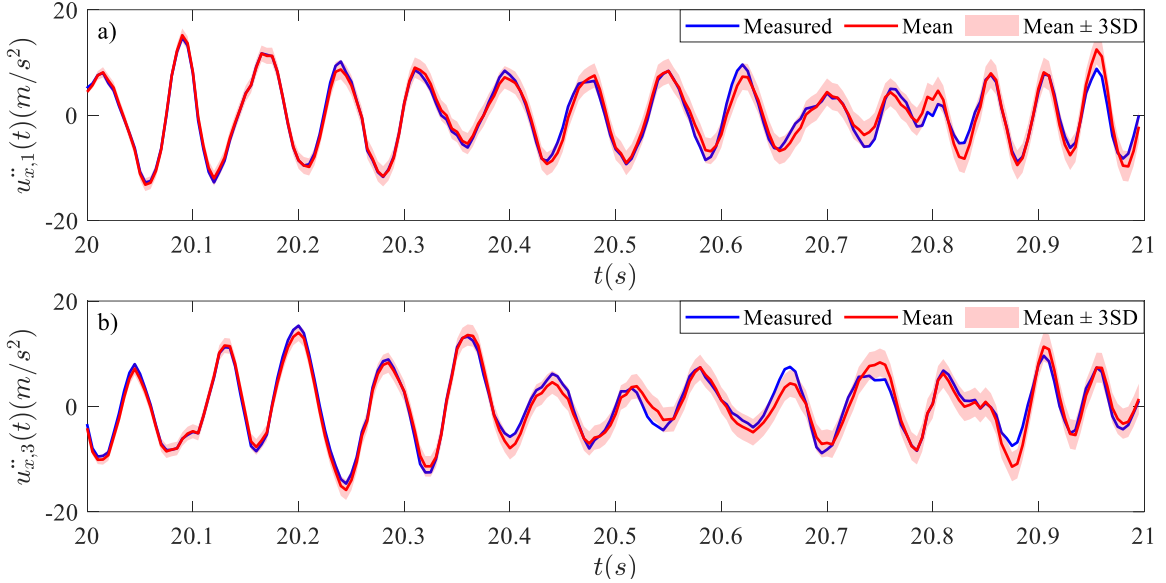
**Fig. 13.** Prediction of the acceleration response of the 1st and 3rd floors based on $p(\mathbf{y}_{N_D+1}|\mathbf{X},\mathbf{Y})$

## 6. Conclusions and Future Works

New kernel covariance functions are introduced to promote physics-informed GP models in the context of structural dynamics inverse problems. These covariance functions are composed of the MMTE kernel function, a physics-driven tangent covariance function, and an isotropic Gaussian noise. When this composite kernel function is studied theoretically, we realize that it incorporates the gradients of physical responses into the covariance structure, offering additional robustness when facing parameter variation. Moreover, hierarchical Bayesian modeling techniques are employed to account for temporal variability by partitioning the measured time-history data into multiple sub-segments. Although the primary motivation behind using a hierarchical probabilistic model is to consider the non-stationarity of measured responses, it turns out that the proposed data partitioning is advantageous from a computational standpoint, where the scalability of GPs can be alleviated. Subsequently, a computational algorithm is proposed to facilitate the sequential processing of measured data and estimate structural parameters and unmeasured dynamical responses. Two illustrative examples are finally presented to validate the proposed methodology, which helps draw the following specific conclusions:

- The spectral density of the model discrepancy reveals the existence of a few dominant peaks. The MMTE kernel function is a suitable kernel function to characterize this stochastic behavior.
- The order of the MMTE kernel function is selected through a Bayesian model selection approach, revealing that a good model order for the MMTE kernel can be the number of dominant peaks appearing on the PSD of the prediction errors.
- The proposed framework reliably accounts for the variability of structural model parameters across the datasets and provides realistic posterior distributions for unmeasured dynamical responses. Unlike other non-hierarchical approaches, the uncertainties calculated do not decrease with the number of data points and partitions.
- The predicted responses show a very reasonable agreement with the measurements, and the propagated uncertainty can contain potential discrepancies of mechanics-based models accounting for possible modeling errors.

Although simple examples were provided to demonstrate the method, this study reveals that higher accuracy can be achieved by incorporating physical knowledge into the kernel covariance function. Nonetheless, more research should be conducted to further discover the applicability of the presented framework to other types of problems. In a broader context, the proposed framework can be explored further by applying it to non-linear systems and non-stationary types of force excitations.

**Appendix (A).** Derivation of the explicit integral in Eq. (8)

**Theorem 1.** Let $\mathbf{X} \in \mathbb{R}^{N_x}$ and $\boldsymbol{\mu} \in \mathbb{R}^{N_\mu}$ be two vectors of random variables described by Gaussian distributions $\mathbf{X} \sim N(\mathbf{X}|\mathbf{A}\boldsymbol{\mu},\boldsymbol{\Sigma})$ and $\boldsymbol{\mu} \sim N(\boldsymbol{\mu}|\boldsymbol{\mu}_0,\boldsymbol{\Sigma}_0)$, where $\mathbf{A} \in \mathbb{R}^{N_x \times N_\mu}$, $\boldsymbol{\mu}_0 \in \mathbb{R}^{N_\mu}$, and $\boldsymbol{\Sigma}_0 \in \mathbb{R}^{N_\mu \times N_\mu}$ are known and given. The integral of the multiplication of these two Gaussian distributions over $\boldsymbol{\mu}$ is given as (See Appendix A of [46] for its detailed proof)

$$\int_{\boldsymbol{\mu}} N(\mathbf{X}|\mathbf{A}\boldsymbol{\mu},\boldsymbol{\Sigma})N(\boldsymbol{\mu}|\boldsymbol{\mu}_0,\boldsymbol{\Sigma}_0)d\boldsymbol{\mu} = N(\mathbf{X}|\mathbf{A}\boldsymbol{\mu}_0,\boldsymbol{\Sigma} + \mathbf{A}\boldsymbol{\Sigma}_0\mathbf{A}^T) \tag{A1}$$

**Corollary 1.** Based on Theorem 1, the integral in Eq. (8) can be calculated as

$$\int_{\boldsymbol{\theta}} N(\mathbf{y}|f(\mathbf{x},\boldsymbol{\theta}),\mathbf{K})N(\boldsymbol{\theta}|\boldsymbol{\mu}_{\boldsymbol{\theta}},\boldsymbol{\Sigma}_{\boldsymbol{\theta}})d\boldsymbol{\theta}$$
$$\approx \int_{\boldsymbol{\theta}} N(\mathbf{y}|f(\mathbf{x},\boldsymbol{\mu}_{\boldsymbol{\theta}}) + \mathbf{J}(\boldsymbol{\theta}-\boldsymbol{\mu}_{\boldsymbol{\theta}}),\mathbf{K})N(\boldsymbol{\theta}|\boldsymbol{\mu}_{\boldsymbol{\theta}},\boldsymbol{\Sigma}_{\boldsymbol{\theta}})d\boldsymbol{\theta} \tag{A2}$$
$$= N(\mathbf{y}|f(\mathbf{x},\boldsymbol{\mu}_{\boldsymbol{\theta}}),\mathbf{J}\boldsymbol{\Sigma}_{\boldsymbol{\theta}}\mathbf{J}^T + \mathbf{K})$$

This finding proves the result claimed earlier in Eq. (8).

**Appendix (B).** Derivation of Eq. (13)

By definition, the PSD of the kernel covariance function given in Eq. (12) can be calculated as

$$S_\varepsilon(\omega;\boldsymbol{\phi}) \triangleq \int_{-\infty}^{\infty} \exp(-j\omega\tau)k_\varepsilon(\tau;\boldsymbol{\phi})d\tau$$
$$= \sum_{k=1}^{m}\left[\int_{-\infty}^{\infty}\sigma_k^2 \exp(-j\omega\tau)\exp(-\tau^2/\ell_k^2)\cos(\omega_k\tau)d\tau\right] + \sigma_n^2\int_{-\infty}^{\infty}\exp(-j\omega\tau)\delta(\tau)d\tau \tag{B1}$$

Note that the integral $\int_{-\infty}^{\infty}\exp(-j\omega\tau)\delta(\tau)d\tau$ is equal to one, due to the underlying properties of Delta function. Additionally, the Euler's formula allows writing the cosine function as $\cos(\omega_{k,i}\tau) = (\exp(j\omega\tau) + \exp(-j\omega\tau))/2$. Replacing this expression into Eq. (B1) yields:

$$S_\varepsilon(\omega;\boldsymbol{\phi}) = \frac{1}{2}\sum_{k=1}^{m}\sigma_k^2\left[\int_{-\infty}^{\infty}\exp\left(-\frac{\tau^2}{\ell_k^2}-(j\omega+j\omega_k)\tau\right) + \exp\left(-\frac{\tau^2}{\ell_k^2}-(j\omega-j\omega_k)\tau\right)d\tau\right] + \sigma_n^2$$
$$= \frac{1}{2}\sum_{k=1}^{m}\sigma_k^2\exp\left(-\frac{\ell_k^2}{4}(\omega+\omega_k)^2\right)\left[\int_{-\infty}^{\infty}\exp\left(-\left(\frac{\tau}{\ell_k}+\frac{\ell_k}{2}(j\omega+j\omega_k)\right)^2\right)d\tau\right] \tag{B2}$$
$$+ \frac{1}{2}\sum_{k=1}^{m}\sigma_k^2\exp\left(-\frac{\ell_k^2}{4}(\omega-\omega_k)^2\right)\left[\int_{-\infty}^{\infty}\exp\left(-\left(\frac{\tau}{\ell_k}+\frac{\ell_k}{2}(j\omega-j\omega_k)\right)^2\right)d\tau\right] + \sigma_n^2$$

The integral $\int_{-\infty}^{\infty} e^{-(ax+b)^2}dx$ is known to be $\sqrt{\pi}/a$, so the latest integrals can be simplified into

$$S_\varepsilon(\omega;\boldsymbol{\phi}) = \frac{1}{2}\sum_{k=1}^{m}\sqrt{\pi}\sigma_k^2\ell_k\left[\exp\left(-\frac{\ell_k^2}{4}(\omega+\omega_k)^2\right) + \exp\left(-\frac{\ell_k^2}{4}(\omega-\omega_k)^2\right)\right] + \sigma_n^2 \tag{B3}$$

This result proves Eq. (13), claimed earlier within the text.

**Appendix (C).** Derivation of Eqs. (30-31)

**Theorem 2.** Let $\mathbf{X} \in \mathbb{R}^{N_x}$ be a vector of random variables described by $N(\mathbf{X}|\boldsymbol{\mu},\boldsymbol{\Sigma})$, where the mean $\boldsymbol{\mu} \in \mathbb{R}^{N_x}$ and covariance $\boldsymbol{\Sigma} \in \mathbb{R}^{N_x \times N_x}$ are both known. For partitioned vector $\mathbf{X} = [\mathbf{x}_1^T \ \mathbf{x}_2^T]^T$, the probability distribution $N(\mathbf{X}|\boldsymbol{\mu},\boldsymbol{\Sigma})$ can be written as

$$p(\mathbf{X}) = N\left(\underbrace{\begin{bmatrix}\mathbf{x}_1\\\mathbf{x}_2\end{bmatrix}}_{\mathbf{X}} \Bigg| \underbrace{\begin{bmatrix}\boldsymbol{\mu}_1\\\boldsymbol{\mu}_2\end{bmatrix}}_{\boldsymbol{\mu}}, \underbrace{\begin{bmatrix}\boldsymbol{\Sigma}_{11} & \boldsymbol{\Sigma}_{12}\\\boldsymbol{\Sigma}_{12}^T & \boldsymbol{\Sigma}_{22}\end{bmatrix}}_{\boldsymbol{\Sigma}}\right) \tag{C1}$$

Then, the probability distribution of $\mathbf{x}_2$ conditional on $\mathbf{x}_1$ can be calculated from [36]:

$$p(\mathbf{x}_2|\mathbf{x}_1) = \frac{p(\mathbf{x}_1,\mathbf{x}_2)}{p(\mathbf{x}_1)} = N(\mathbf{x}_2|\boldsymbol{\mu}_{2|1},\boldsymbol{\Sigma}_{2|1}) \tag{C2}$$

$$\boldsymbol{\mu}_{2|1} = \boldsymbol{\mu}_2 + \boldsymbol{\Sigma}_{12}^T \boldsymbol{\Sigma}_{11}^{-1}(\mathbf{x}_1 - \boldsymbol{\mu}_1) \tag{C3}$$

$$\boldsymbol{\Sigma}_{2|1} = \boldsymbol{\Sigma}_{22} - \boldsymbol{\Sigma}_{12}^T \boldsymbol{\Sigma}_{11}^{-1} \boldsymbol{\Sigma}_{12} \tag{C4}$$

**Acknowledgements**
Financial support from the Hong Kong Research Grants Council under project numbers 16212918 and 16211019 is gratefully appreciated.

**References**
[1]  J.L. Beck, L.S. Katafygiotis, Updating models and their uncertainties. I: Bayesian statistical framework, J. Eng. Mech. 124 (1998) 455–461. https://doi.org/10.1061/(ASCE)0733-9399(1998)124:4(455).
[2]  L.S. Katafygiotis, J.L. Beck, Updating models and their uncertainties. II: model identifiability, J. Eng. Mech. 124 (1998) 463–467. https://doi.org/10.1061/(ASCE)0733-9399(1998)124:4(463).
[3]  M. Raissi, P. Perdikaris, G.E. Karniadakis, Physics-informed neural networks: A deep learning framework for solving forward and inverse problems involving nonlinear partial differential equations, J. Comput. Phys. 378 (2019) 686–707. https://doi.org/10.1016/j.jcp.2018.10.045.
[4]  M. Raissi, P. Perdikaris, G.E. Karniadakis, Numerical Gaussian Processes for Time-Dependent and Nonlinear Partial Differential Equations, SIAM J. Sci. Comput. 40 (2018) A172–A198. https://doi.org/10.1137/17M1120762.
[5]  R.M. Neal, Bayesian Learning for Neural Networks, Springer New York, New York, NY, 1996. https://doi.org/10.1007/978-1-4612-0745-0.
[6]  A. Jacot, F. Gabriel, C. Hongler, Neural Tangent Kernel: Convergence and Generalization in Neural Networks, (2018). http://arxiv.org/abs/1806.07572.
[7]  C.E. Rasmussen, C.K.I. Williams, Gaussian Processes for Machine Learning, MIT Press, 2005.
[8]  K.P. Murphy, Machine learning: a probabilistic perspective, MIT Press, 2012. https://doi.org/10.1007/978-94-011-3532-0_2.
[9]  M.C. Kennedy, A. O'Hagan, Bayesian calibration of computer models, J. R. Stat. Soc. Ser. B Stat. Methodol. 63 (2001) 425–464. https://doi.org/10.1111/1467-9868.00294.
[10] E. Simoen, C. Papadimitriou, G. Lombaert, On prediction error correlation in Bayesian model updating, J. Sound Vib. 332 (2013) 4136–4152. https://doi.org/10.1016/j.jsv.2013.03.019.
[11] I. Bilionis, N. Zabaras, B.A. Konomi, G. Lin, Multi-output separable Gaussian process: Towards an efficient, fully Bayesian paradigm for uncertainty quantification, J. Comput.

Phys. 241 (2013) 212–239. https://doi.org/10.1016/j.jcp.2013.01.011.

[12] C.M. Bishop, Pattern Recognition and Machine Learning, New York, 2006.

[13] G. Jia, A.A. Taflanidis, Kriging metamodeling for approximation of high-dimensional wave and surge responses in real-time storm/hurricane risk assessment, Comput. Methods Appl. Mech. Eng. 261–262 (2013) 24–38. https://doi.org/10.1016/j.cma.2013.03.012.

[14] P. Angelikopoulos, C. Papadimitriou, P. Koumoutsakos, X-TMCMC: Adaptive kriging for Bayesian inverse modeling, Comput. Methods Appl. Mech. Eng. 289 (2015) 409–428.

[15] H.A. Jensen, C. Esse, V. Araya, C. Papadimitriou, Implementation of an adaptive meta-model for Bayesian finite element model updating in time domain, Reliab. Eng. Syst. Saf. 160 (2017) 174–190. https://doi.org/10.1016/j.ress.2016.12.005.

[16] A.M. Kosikova, O. Sedehi, L.S. Katafygiotis, Bayesian Model Updating using Gaussian Process Regression, in: J. Li, P.D. Spanos, J.B. Chen, Y.B. Peng (Eds.), 13th Int. Conf. Struct. Saf. Reliab. (ICOSSAR 2021), Shanghai, China, 2021.

[17] K.V. Yuen, Bayesian Methods for Structural Dynamics and Civil Engineering, 2010. https://doi.org/10.1002/9780470824566.

[18] C.R. Farrar, K. Worden, Structural health monitoring: a machine learning perspective, John Wiley & Sons, 2012.

[19] S.-K. Au, Operational Modal Analysis-Modeling, Bayesian Inference, Uncertainty Laws, Springer, Singapore, 2017. https://doi.org/10.1007/978-981-10-4118-1.

[20] K.-V. Yuen, L.S. Katafygiotis, Bayesian time–domain approach for modal updating using ambient data, Probabilistic Eng. Mech. 16 (2001) 219–231. https://doi.org/10.1016/S0266-8920(01)00004-2.

[21] K.-V. Yuen, L.S. Katafygiotis, Bayesian Modal Updating using Complete Input and Incomplete Response Noisy Measurements, J. Eng. Mech. 128 (2002) 340–350. https://doi.org/10.1061/(ASCE)0733-9399(2002)128:3(340).

[22] C. Papadimitriou, G. Lombaert, The effect of prediction error correlation on optimal sensor placement in structural dynamics, Mech. Syst. Signal Process. 28 (2012) 105–127. https://doi.org/10.1016/j.ymssp.2011.05.019.

[23] L.D. Avendaño-Valencia, E.N. Chatzi, D. Tcherniak, Gaussian process models for mitigation of operational variability in the structural health monitoring of wind turbines, Mech. Syst. Signal Process. 142 (2020) 106686. https://doi.org/10.1016/j.ymssp.2020.106686.

[24] Y.-C. Zhu, S.-K. Au, Bayesian data driven model for uncertain modal properties identified from operational modal analysis, Mech. Syst. Signal Process. 136 (2020) 106511. https://doi.org/10.1016/j.ymssp.2019.106511.

[25] R. Nayek, S. Chakraborty, S. Narasimhan, A Gaussian process latent force model for joint input-state estimation in linear structural systems, Mech. Syst. Signal Process. 128 (2019) 497–530. https://doi.org/10.1016/j.ymssp.2019.03.048.

[26] T.J. Rogers, K. Worden, E.J. Cross, On the application of Gaussian process latent force models for joint input-state-parameter estimation: With a view to Bayesian operational identification, Mech. Syst. Signal Process. 140 (2020) 106580. https://doi.org/10.1016/j.ymssp.2019.106580.

[27] C. Jiang, M.A. Vega, M.K. Ramancha, M.D. Todd, J.P. Conte, M. Parno, Z. Hu, Bayesian calibration of multi-level model with unobservable distributed response and application to miter gates, Mech. Syst. Signal Process. 170 (2022) 108852. https://doi.org/10.1016/j.ymssp.2022.108852.

[28] P. Gardner, L.A. Bull, N. Dervilis, K. Worden, On the application of kernelised Bayesian transfer learning to population-based structural health monitoring, Mech. Syst. Signal Process. 167 (2022) 108519. https://doi.org/10.1016/j.ymssp.2021.108519.

[29] M.K. Ramancha, J.P. Conte, M.D. Parno, Accounting for model form uncertainty in Bayesian calibration of linear dynamic systems, Mech. Syst. Signal Process. 171 (2022) 108871. https://doi.org/10.1016/j.ymssp.2022.108871.

[30] M.R. Jones, T.J. Rogers, E.J. Cross, Constraining Gaussian processes for physics-informed acoustic emission mapping, Mech. Syst. Signal Process. 188 (2023) 109984. https://doi.org/10.1016/j.ymssp.2022.109984.

[31] O. Sedehi, C. Papadimitriou, L.S. Katafygiotis, Probabilistic hierarchical Bayesian framework for time-domain model updating and robust predictions, Mech. Syst. Signal Process. 123 (2019) 648–673. https://doi.org/10.1016/j.ymssp.2018.09.041.

[32] M. Song, I. Behmanesh, B. Moaveni, C. Papadimitriou, Accounting for modeling errors and inherent structural variability through a hierarchical bayesian model updating approach: An overview, Sensors (Switzerland). 20 (2020) 1–27. https://doi.org/10.3390/s20143874.

[33] O. Sedehi, C. Papadimitriou, L.S. Katafygiotis, Hierarchical Bayesian uncertainty quantification of Finite Element models using modal statistical information, Mech. Syst. Signal Process. 179 (2022) 109296. https://doi.org/10.1016/j.ymssp.2022.109296.

[34] I. Behmanesh, B. Moaveni, G. Lombaert, C. Papadimitriou, Hierarchical Bayesian model updating for structural identification, Mech. Syst. Signal Process. 64–65 (2015) 360–376. https://doi.org/10.1016/j.ymssp.2015.03.026.

[35] K.-V. Yuen, L.S. Katafygiotis, Bayesian time–domain approach for modal updating using ambient data, Probabilistic Eng. Mech. 16 (2001) 219–231. https://doi.org/10.1016/S0266-8920(01)00004-2.

[36] K.B. Petersen, M.S. Pedersen, The Matrix Cookbook, Technical University of Denmark, 2012.

[37] M.I. Jordan, Learning in graphical models, Springer Science & Business Media, 1998.

[38] J.B. Nagel, B. Sudret, A unified framework for multilevel uncertainty quantification in Bayesian inverse problems, Probabilistic Eng. Mech. 43 (2016) 68–84. https://doi.org/10.1016/j.probengmech.2015.09.007.

[39] J. Quiñonero-Candela, C.E. Rasmussen, A unifying view of sparse approximate Gaussian process regression, J. Mach. Learn. Res. 6 (2005) 1939–1959.

[40] E.A. Johnson, H.F. Lam, L.S. Katafygiotis, J.L. Beck, Phase I IASC-ASCE Structural Health Monitoring Benchmark Problem Using Simulated Data, J. Eng. Mech. 130 (2004) 3–15. https://doi.org/10.1061/(ASCE)0733-9399(2004)130:1(3).

[41] O. Sedehi, C. Papadimitriou, D. Teymouri, L.S. Katafygiotis, Sequential Bayesian estimation of state and input in dynamical systems using output-only measurements, Mech. Syst. Signal Process. 131 (2019) 659–688. https://doi.org/10.1016/j.ymssp.2019.06.007.

[42] D. Teymouri, O. Sedehi, L.S. Katafygiotis, C. Papadimitriou, Input-state-parameter-noise identification and virtual sensing in dynamical systems: A Bayesian expectation-maximization (BEM) perspective, Mech. Syst. Signal Process. 185 (2023) 109758. https://doi.org/10.1016/j.ymssp.2022.109758.

[43] J. Ching, Y.-C. Chen, Transitional Markov Chain Monte Carlo Method for Bayesian Model Updating, Model Class Selection, and Model Averaging, J. Eng. Mech. 133 (2007) 816–832. https://doi.org/10.1061/(ASCE)0733-9399(2007)133:7(816).

[44] S. Wu, P. Angelikopoulos, C. Papadimitriou, P. Koumoutsakos, Bayesian Annealed Sequential Importance Sampling: An Unbiased Version of Transitional Markov Chain Monte Carlo, ASCE-ASME J. Risk Uncert. Engrg. Sys., Part B Mech. Engrg. 4 (2017) 011008. https://doi.org/10.1115/1.4037450.

[45] W. Betz, I. Papaioannou, D. Straub, Transitional Markov Chain Monte Carlo: Observations and Improvements, J. Eng. Mech. 142 (2016) 04016016. https://doi.org/10.1061/(ASCE)EM.1943-7889.0001066.

[46] O. Sedehi, L.S. Katafygiotis, C. Papadimitriou, Hierarchical Bayesian operational modal analysis: Theory and computations, Mech. Syst. Signal Process. 140 (2020). https://doi.org/10.1016/j.ymssp.2020.106663.